\documentclass[journal]{IEEEtran}
\usepackage{color}
\usepackage{yfonts}
\usepackage{cite}
\usepackage[pdftex]{graphicx}
\setkeys{Gin}{clip=true,draft=false}
\DeclareGraphicsExtensions{.pdf}
\usepackage[cmex10]{amsmath}
\usepackage{amsthm}
\usepackage{amsfonts}
\usepackage{amssymb}
\usepackage[cspex,bbgreekl]{mathbbol}
\usepackage[hang]{subfigure}
\usepackage{color}
\usepackage{float}
\usepackage[super]{nth}
\newcommand{\xvec}{\mathbf{x}}
\newcommand{\yvec}{\mathbf{y}}
\newcommand{\zvec}{\mathbf{z}}
\newcommand{\svec}{\mathbf{\mathfrak{s}}}

\begin{document}

\title{Universal Memcomputing Machines}



\author{Fabio L. Traversa, Massimiliano Di Ventra\thanks{The authors are with the Department of Physics, University of California-San Diego, 9500  Gilman Drive, La Jolla, California 92093-0319, USA, e-mail: fabio.traversa@polito.it, diventra@physics.ucsd.edu}}

\maketitle

\date{\today}
\begin{abstract}
We introduce the notion of {\it universal memcomputing machines} (UMMs): a class of brain-inspired general-purpose computing machines based on systems with memory, whereby processing and storing of information occur on the same physical location. We analytically prove that the memory properties of UMMs endow them with {\it universal computing power}---they are Turing-complete---, {\it intrinsic parallelism}, {\it functional polymorphism}, and {\it information overhead}, namely their collective states can support  exponential data compression directly in memory. We also demonstrate that a UMM has the same computational power as a non-deterministic Turing machine, namely it can solve NP--complete problems in polynomial time. However, by virtue of its information overhead, a UMM needs only an amount of memory cells (memprocessors) that grows polynomially with the problem size. As an example we provide the polynomial-time solution of the subset-sum problem and a simple hardware implementation of the same. Even though these results do not prove the statement NP=P within the Turing paradigm, the practical realization of these UMMs would represent a paradigm shift from present von Neumann architectures bringing us closer to brain-like neural computation.
\end{abstract}



\section{Introduction}\label{introduction}

\begin{figure*}
\centerline{
\includegraphics[width=1.8\columnwidth]{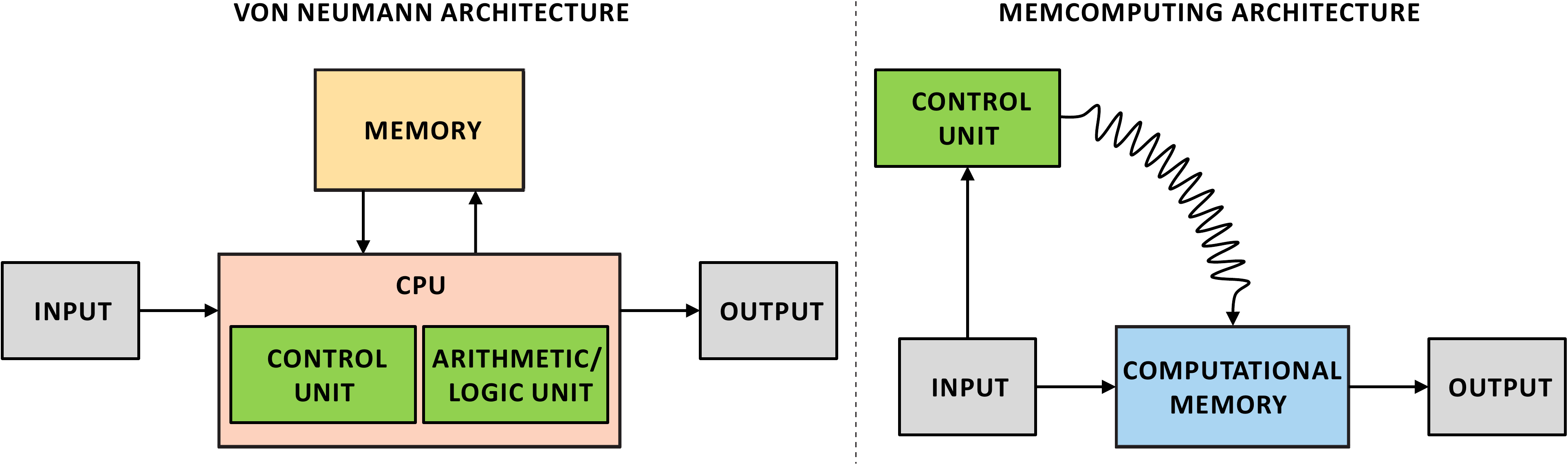}}
\caption{\label{architecture}Von Neumann vs. memcomputing architecture. The straight arrows indicate
the flow of data, while the wiggly arrow indicates only that a signal is being sent. In the memcomputing
architecture the input data can either feed directly into the memprocessors composing the computational memory, or are fed into the control unit.}
\end{figure*}

Since Alan Turing invented his ideal machine in 1936 \cite{36_turing,turing_book}, mathematicians have been able to develop this concept into what is now known as computational complexity theory \cite{computational_complexity_book}, a powerful tool essentially employed to determine how long does an algorithm take to solve a problem with given input data. This ideal machine is now known as universal Turing machine (UTM) and represents the conceptual underpinning of all our modern-day digital computers.

The practical realization of a UTM is commonly done using the von Neumann architecture \cite{VNA}, which apart from some inconsequential details, it can be viewed as a device that requires a central processing unit (CPU) that is {\it physically separate} from the memory (see figure \ref{architecture}, left panel). The CPU contains both a {\it control unit} that directs the operation of the machine, and {\it all} arithmetic functions and logic gates the machine needs during execution ({\it arithmetic/logic unit}). This form of computation requires a large amount of data to be transferred between the CPU and the memory, and this necessarily limits the machine both in time (von Neumann bottleneck \cite{78_Backus}), and in energy~\cite{computer_architecture_book}.

Parallel computation, as it is currently envisioned, mitigates somewhat, but does not solve any of these issues: several processors each manipulate parts of the whole data, by typically working with a physically ``close'' memory. Eventually, however, all the processors have to communicate with each other to solve the whole problem, still requiring a substantial amount of information transfer between them and their memories \cite{computer_architecture_book}. Overcoming this ``information latency issue'' would then require a fundamentally different way in which data are manipulated and stored.

Historically, the first alternative to the von Neumann architecture was the ``Harvard architecture'' developed by H. Aiken, concurrently with the ENIAC project that would be using the von Neumann architecture \cite{computer_architecture_book}. Original Harvard architectures had separate memories for instructions and data. However, this term is used today to mean machines with a single main memory, but with separate instruction and data caches \cite{computer_architecture_book}. A more recent alternative is the ``pipelined architecture'', i.e., data processing stages connected in series, where the output of one stage is the input of the next one \cite{computer_architecture_book}. This architecture is commonly used in modern CPUs, and is particularly efficient for graphics processing units (GPUs) \cite{GPU}. Finally, we mention an alternative concept that is still under investigation, named ``liquid-state machine''. It consists of a computational model for real-time computing on time-varying input \cite{liquid_machine}. Although some of these concepts have found (or may eventually find) use in practical computing, none of these alternatives completely solve the limitations of the von Neumann architecture, or show additional, substantial advantages compared with Turing machines.

Very recently, a new computing paradigm, inspired by the operation of our own brain, has been proposed which is not based on the UTM concept, and which puts the whole burden of computation directly into the memory. This paradigm has been named {\it memcomputing} \cite{13_memcomputing}.

Like the brain, memcomputing machines would compute with and in memory without the need of a separate CPU. The memory allows learning and adaptive capabilities \cite{09_amoeba,13_amoeba}, bypassing broken connections and self-organizing the computation into the solution path \cite{11_maze,13_self_organization}, much like the brain is able to sustain a certain amount of damage and still operate seamlessly.

The whole concept of memcomputing can be realized in practice by utilizing physical properties of many materials and systems which show a certain degree of time non-locality (memory) in their response functions at particular frequencies and strengths of the inputs \cite{09_memory_materials,11_memory_materials,13_properties}. In fact, the field has seen a surge of activities since the introduction of resistors, capacitors, and inductors with memory, (memristors, memcapacitors, meminductors, respectively) collectively called {\it memelements} \cite{09_memelements}, and their actual realization in a variety of systems (see, e.g., \cite{08_strukov,07_waser,09_lu,12_grollier,13_sillin} and the review~\cite{11_memory_materials}). For instance, physical and statistical properties, \cite{Pershin_13,Oskoee_11,Shihong_12,Budhathoki_14} computing capability \cite{borghetti_10,12_Proc_IEEE}, and more \cite{adamatzky_13} have been studied for networks of memristors. However, more complex dynamics and very interesting properties can arise using memelements other than memristors, and combining these with standard electronic devices. For example, we have recently suggested an actual implementation of a memcomputing machine that has the same architecture of a Dynamic Random Access Memory (DRAM) \cite{computer_architecture_book}, but employs memcapacitors to compute and store information \cite{DCRAM}. This particular architecture, we have called Dynamic Computing Random Access Memory (DCRAM) and expends a very small amount of energy per operation, is just an example of what can be accomplished with memelements. Nevertheless, it already shows two features that are not available in our modern computers: {\it intrinsic parallelism} and {\it functional polymorphism}.

\begin{figure}
\centerline{
\includegraphics[width=\columnwidth]{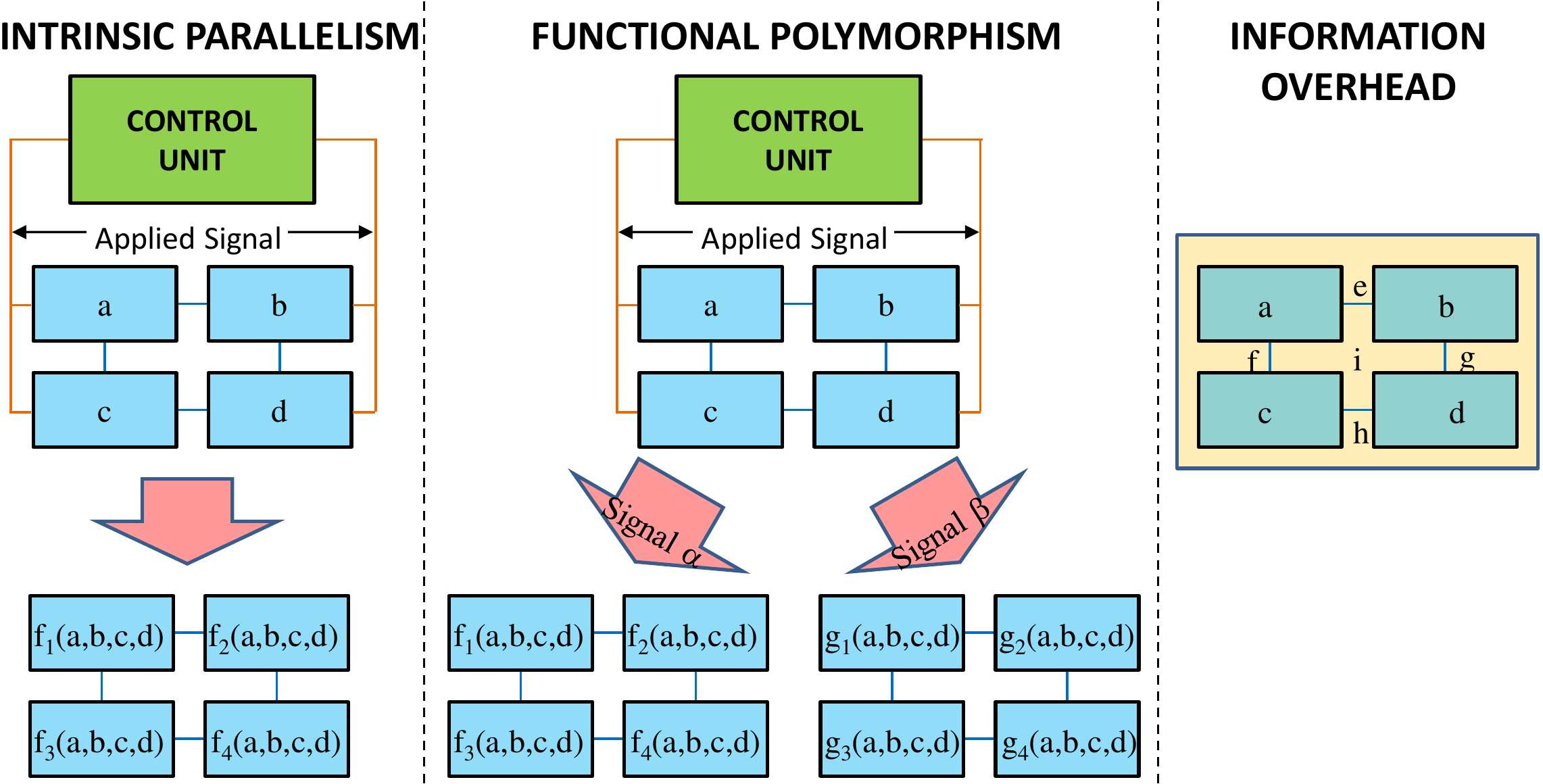}}
\caption{\label{UMM_features}Sketch of the three main features of a UMM: intrinsic parallelism (left panel),
functional polymorphism (middle panel), and information overhead (right panel).}
\end{figure}

The first feature means that {\it all} memelement processors (memory cells or {\it memprocessors} for short), operate {\it simultaneously} and {\it collectively} during the computation (Fig.~\ref{UMM_features} leftmost panel). This way, problems that otherwise would require several steps to be solved can be executed in one or a few steps \cite{11_maze,13_self_organization,09_chua,12_Proc_IEEE}. The second feature relates to the ability of computing different functions without modifying the topology of the machine network, by simply applying the appropriate input signals \cite{DCRAM} (Fig.~\ref{UMM_features} middle panel). This polymorphism, which is again similar to that boasted by our brain and some of our organs (such as the eyes), shows also another important feature of memcomputing machines: their control unit---which is {\it not} a full-fledged CPU since it does not need any arithmetic/logic function, and which is required in order to control the execution of the type of problem that needs to be solved by the machine---can either be fed directly by the input data, or indirectly through the flow of input data to the memprocessors (see figure \ref{architecture}, right panel).

The last important property of UMMs, namely their {\it information overhead}, is related to the way in which memprocessors
{\it interconnected by a physical coupling} can store a certain amount of data (Fig.~\ref{UMM_features} rightmost panel). More specifically, the information overhead is the capability of an interacting memprocessor network to store and compress an information amount larger than that possible by the same non-interacting memprocessors. In fact, in section \ref{information_overhead} we will show how a linear number of interconnected memprocessors can store and compress an amount of data that grows even exponentially. It is this {\it physical} type of interaction--resembling closely that of the neurons in the brain--which also underlies a fundamental difference with the present UTMs and their practical realizations.

Having found an example of a memcomputing machine and its main features, it is then natural to ask whether, in analogy with a general-purpose UTM, a similar concept can be defined in the case of memcomputing machines. And if so, how do they relate to UTMs and what types of computational complexity classes these systems would be able to solve?

In this paper, we introduce the notion of {\it universal memcomputing machine} (UMM). This is a class of general-purpose machines based on interacting memprocessors. In the present context, we do not limit this concept to memprocessors composed only by memristors, memcapacitors or meminductors. Rather, we consider any system whose response function to some external stimuli shows some degree of memory~\cite{11_memory_materials,13_properties}. We show that such an ideal machine is Turing-complete, namely it can
simulate a UTM. Most importantly, however, in view of its intrinsic parallelism, functional polymorphism and information overhead, we prove that a UMM is able to solve Non-deterministic Polynomial (NP) problems in polynomial (P) time. In particular, we consider the NP-complete subset-sum problem, and using two examples we show how a UMM can solve it in polynomial time using a number of memprocessors that either grows exponentially or, by exploiting the information overhead, uses a linear number of memprocessors. This last example can be easily realized in an electronics lab.

Note that this result does {\it not} prove that NP=P, since this proposition should be solved only within the domain of UTMs. Rather, it shows that the concept we introduce, if realized in practice, provides a powerful new paradigm for computation with features approaching those of a brain.

The paper is organized as follows. In Sec.~\ref{UTM} we first briefly review the formal definition of
UTMs. In Sec.~\ref{memprocessor} we introduce the formal definition of memprocessor while in Sec.~\ref{UMMSec} we introduce the mathematical description of a UMM and prove its Turing-completeness. In Sec.~\ref{UMMSecP} we discuss the first two properties of UMMs, specifically their parallelism and polymorphism. We devote the entire Sec.~\ref{information_overhead} to the third important property of UMMs, namely their information overhead. We use these properties in Sec.~\ref{SSPSec} to solve the subset-sum problem in P--time. We
finally conclude with our thoughts for the future in Sec.~\ref{Conclusions}.

\section{Universal Turing Machine}\label{UTM}

In order to properly clarify the aim of our present work and put it into context, we briefly review the point of view of the UTM. For a complete and thorough description we refer the reader to the extensive literature on the subject (\cite{turing_book} for history on UTM, \cite{complexity_bible} a classic, and \cite{computational_complexity_book} for more modern literature).

In simple terms a UTM is a mathematical model of an object operating on a tape. The fundamental ingredients of a UTM are then the \textit{tape} itself (possibly infinite), a \textit{tape head} to read and write on the tape, a \textit{state register} storing the states of the UTM (including the initial and final states),  and finally a finite \textit{table} of instructions.
Formally, a UTM is defined as the seven-tuple
\begin{equation}
UTM=(Q,\Gamma,b,\Sigma,\delta,q_0,F)\,,\label{UTMdef}
\end{equation}
where $Q$ is the set of possible states that the UTM can be in, $\Gamma$
is the set of symbols that can be written on the tape, $b\in\Gamma$ is the blank symbol, $\Sigma$ is the set of input symbols, $q_0\in Q$
is the initial state of the UTM, $F\subseteq Q$ is the set of final states, and
\begin{equation}
\delta:Q\backslash F\times\Gamma\rightarrow Q\times\Gamma\times\{L,N,R\}\,,\label{deltaUTM}
\end{equation}
is the transition function (the table) where $L$, $N$ and $R$ are Left, No, and Right shift, respectively.

From this formal definition it can be readily seen that no explicit information about any physical system or its properties is involved in a UTM. In other words, the UTM is just an abstract mathematical object, through which it is possible to determine the computational complexity of an algorithm, i.e., given $n=\text{cardinality}[\Sigma]$ we can determine the number of operations $f(n,\delta)$ the UTM should perform in order to find the final state $F$.

In the last sixty years, several alternatives or modifications of the UTM have been introduced, from the non-deterministic Turing machine to parallel Turing machine and more, but all of those variants have been unified through the concept of universality of a UTM \cite{complexity_bible,computational_complexity_book}. In this picture, the complexity theory derived from each of those variants is equivalent, in the sense that all these machines can
solve the same type of problem, albeit with different execution times. A separate discussion is required for the Quantum Turing machine, and
indeed we will compare this to a UMM in Sec.~\ref{UMMSecP}.

\section{Memprocessors}\label{memprocessor}

The basic constitutive unit of a memcomputing architecture is what we name ``memprocessor''. Here, we give a formal definition of memprocessor that actually represents the link between real memcomputing architectures and the formal definition of a UMM. This should also clarify that a memprocessor is not necessarily made of passive elements with memory,
such as memelements, rather it is a much more general object.

We define a memprocessor as an object defined by the four-tuple $(x,y,z,\sigma)$ where $x$ is the state of the memprocessor, $y$ is the array of internal variables, $z$ the array of variables that connect from one
memprocessor to other memprocessors, and $\sigma$ an operator that defines the evolution
\begin{equation}\label{memprocdef}
\sigma[ x,y,z]=(x^{\prime},y^{\prime}).
\end{equation}

When two or more memprocessors are connected, we have a network of memprocessors (\emph{computational memory}). In this case we define the vector $\xvec$ as the state of the network (i.e., the array of all the states $x_{i}$ of each memprocessor), and $\zvec=\cup_iz_i$ the array of all connecting variables, with $z_i$ the connecting array of variables of the memprocessor $i$-th.

Let $z_i$ and $z_j$ be respectively the vectors of connecting variables of the memprocessors $i$ and $j$, then if $z_i\cap z_j\neq\varnothing$ we say that the two memprocessors are connected. Alternatively, a memprocessor is not connected to any other memprocessor (\emph{isolated}) when we have $z=z(x,y)$ (i.e., $z$ is completely determined by $x$ and $y$) and
\begin{equation}
\sigma[ x,y,z(x,y)]=(x,y),
\end{equation}
which means that the memprocessor has no dynamics.

A network of memprocessors has the evolution of the connecting variables $\zvec$ given by the evolution operator $\Xi$ defined as
\begin{equation}
\Xi[\xvec,\yvec,\zvec,\svec]=\zvec\rq{}
\end{equation}
where $\yvec=\cup_iy_i$ and $\svec$ is the array of the external signals that can be applied to a subset of connections to provide stimuli for the network. Finally, the complete evolution of the network is defined by the system
\begin{equation}\label{memprocnetdef}
\left\{\begin{array}{l c l}
\sigma[ x_1,y_1,z_1]&=&(x_1\rq{},y_1\rq{})\\
&\vdots& \\
\sigma[ x_n,y_n,z_n]&=&(x_n\rq{},y_n\rq{})\\
\Xi[\xvec,\yvec,\zvec,\svec]&=&\zvec\rq{}.
\end{array}\right.
\end{equation}

The evolution operators $\sigma$ and $\Xi$ can be interpreted either as discrete or continuous evolution operators. The discrete evolution operator interpretation includes also the artificial neural networks\cite{neural_networks_bible}, while the continuous operator interpretation represents more general dynamical systems. We analyze two types of continuous operators: the operators representing memprocessors made only by memelements, and memprocessors given by arbitrary electronic units.

\subsection{Memprocessors made by memelements}

The standard equations of a memelement are given by the following relations \cite{09_memelements}
\begin{align}
x_{ex}(t)  & =g(x_{in},u,t)u(t)\label{mem1}\\
\dot x_{in}(t)  & =f(x_{in},u,t)\label{mem2}
\end{align}
where $x_{in}$ denotes a set of state variables describing the internal state of the system, $u$ and $x_{ex}$ are any two complementary constitutive variables (i.e., current, charge, voltage, or flux) denoting input and output of the system, $g$ is a generalized response, and $f$ a continuous vector function. In this case the connection variables $z$ are $x_{ex}(t)$ and $u(t)$, i.e.,
$z=[x_{ex}(t),u(t)]$, and the state $x$ coincides with $x_{in}(t)$. Now, \eqref{mem1} can be interpreted as a constraint for \eqref{mem2}, in the sense that we can write $u=u(x_{in}(t),x_{ex}(t))$ solution of $x_{ex}(t)=g(x_{in},u,t)u(t)$,
so we can simply replace $u(x_{in},y)$ in \eqref{mem2}, and forget about \eqref{mem1}. In this case the evolution in a time $T$ of \eqref{mem2} is given by
\begin{equation}
x_{in}(t+T)-x_{in}(t)=\int_t^{t+T}f(x_{in}(\tau),u(x_{in}(\tau),x_{ex}(\tau)),\tau)d\tau
\end{equation}
and defining $x=x_{in}(t)$ and $x\rq{}=x_{in}(t+T)$ we have
\begin{align}
\sigma[ x,y,z]&=\sigma[ x,z]= \nonumber \\
                               =x_{in}(t&)+\int_t^{t+T}f(x_{in}(\tau),u(x_{in}(\tau),x_{ex}(\tau)),\tau)d\tau\,.
\end{align}
On the other hand, the operator $\Xi$ will be simply defined by the Kirchhoff's laws and external generators in order to include the external signals $\svec$.

\subsection{Memprocessors made of generic electronic devices}

In this case, the equation that defines a memprocessor is a general Differential Algebraic Equation (DAE) system that represents any electronic circuit. It can be casted as
\begin{equation}
\frac{d}{dt}q(x,y,z)=f(x,y,z,t)\label{DAE}\,,
\end{equation}
where $(x,y,z)$ represents all the state variables of the circuit.

Using the chain rule we have $\frac{dq}{dt}=\frac{\partial q}{\partial x}\dot x+\frac{\partial q}{\partial y}\dot y+\frac{\partial q}{\partial z}\dot z$. From circuit theory and modified nodal analysis\cite{wiley_enc} there always exists a choice of $y$ such that the Jacobian matrix
\begin{equation}
J_{x,y}(x,y,z)=\left[
\begin{array}
{c c c}%
\cfrac{\partial q(x,y,z)}{\partial x}& &
\cfrac{\partial q(x,y,z)}{\partial y}
\end{array}
\right]\label{jacobian}
\end{equation}
is squared but not necessarily invertible. If the $J_{x,y}$ is not invertible, we can eliminate some variables by including constraints as in the previous section thus obtaining always a (reduced) $J_{x,y}$ which is invertible. Therefore, without loss of generality, we can assume $J_{x,y}$ invertible and we have from the chain rule and the Eqs.~\eqref{DAE} and \eqref{jacobian}
\begin{equation}
\left[
\begin{array}{c}
\dot x\\
\dot y
\end{array}
\right]  =J^{-1}_{x,y}(x,y,z)\left(  f(x,y,z,t)-\frac{\partial q(x,y,z)}{\partial z}\dot z\right)\,.
\end{equation}
The evolution in a time $T$ is then given by
\begin{align}
\sigma&[ x,y,z]=
  \left[\begin{array}{c} x(t+T)\\ y(t+T) \end{array}\right]
=\left[\begin{array}{c} x(t)\\ y(t) \end{array}\right]+\nonumber \\
&+\int_{t}^{t+T}J^{-1}_{x,y}(x,y,z)\left(  f(x,y,z,\tau)-\frac{\partial q(x,y,z)}{\partial z}\dot z\right)   d\tau.
\end{align}

Also in this case the operator $\Xi$ will be simply defined by the Kirchhoff's laws and external generators in order to include the external signals $\svec$. This type of memprocessor represents for example those memory cells (e.g., memcapacitor plus field-effect transistor) of the DCRAM we have introduced \cite{DCRAM}, as well as the memory cells we will describe in Sec.~\ref{hardware_implementation}. However, we anticipate that the latter has the state of the memprocessor defined by the amplitudes of the Fourier series of the state $x$.

\section{Universal Memcomputing Machine}\label{UMMSec}

We are now ready to introduce the general concept of a UMM. The UMM is an ideal machine formed by a bank of interconnected memory cells---memprocessors---(possibly infinite in number) able to perform either digital (logic) or analog (functional) operations controlled by a control unit (see figure~\ref{architecture}, right panel). The computation {\it with} and {\it in} memory can be sketched in the following way. When two or more memprocessors are connected, through a signal sent by the control unit, the memprocessors change their internal states according to both their initial states and the signal, thus giving rise to intrinsic parallelism and the functional polymorphism we have anticipated.

\subsection{Formal definition}\label{formalUMM}

We define the UMM as the eight-tuple
\begin{equation}
UMM=(M,\Delta,{\cal P},S,\Sigma,p_0,s_0,F)\,,\label{UMMdef}
\end{equation}
where $M$ is the set of possible states of a single memprocessor. It can be either a finite set $M_d$ (digital regime), a continuum or an infinite discrete set of states $M_a$ (analog regime), thus $M$ can be expressed as $M=M_d\bigoplus M_a$. $\Delta$ is a set of functions
\begin{equation}
\delta_\alpha:M^{m_\alpha}\backslash F\times {\cal P}\rightarrow M^{m\rq_\alpha}\times {\cal P}^2\times S\,,\label{functUMM}
\end{equation}
where $m_\alpha<\infty$ is the number of memprocessors used as input of (read by) the function $\delta_\alpha$, and $m\rq_\alpha<\infty$ is the number of memprocessors used as output (written by) the function $\delta_\alpha$; ${\cal P}$ is the set of the arrays of pointers $p_\alpha$ that select the memprocessors called by $\delta_\alpha$ and $S$ is the set of indexes $\alpha$; $\Sigma$ is the set of the initial states written by the input device on the computational memory; $p_0\in {\cal P}$ is the initial array of pointers; $s_0$ is the initial index $\alpha$ and $F\subseteq M$ is the set of final states.

For the sake of clarity, we spend a few more words on the functions $\delta_\alpha$. Let $p_\alpha,p\rq_\alpha,p_\beta\in {\cal P}$ be the arrays $p_\alpha=\{i_1,...,i_{m_\alpha} \}$, $p\rq_\alpha=\{j_1,...,j_{m\rq_\alpha} \}$ and $p_\beta \{k_1,...,k_{m_\beta} \}$ and $\beta\in s$, then the function $\delta_\alpha$ can be expressed as
\begin{equation}
\delta_\alpha[\xvec(p_\alpha)]=(\xvec\rq(p\rq_\alpha),\beta,p_\beta)\,,
\label{delta}
\end{equation}
where $\xvec$ is the vector of the states of the memprocessors, thus $\xvec(p_\alpha)\in M^{m_\alpha}$ are the states of the memprocessors selected as input for $\delta_\alpha$, while $\xvec\rq(p\rq_\alpha)\in M^{m\rq_\alpha}$ are the output states of $\delta_\alpha$. Then $\delta_\alpha$ reads the states $\xvec(p_\alpha)$ and writes the new states $\xvec\rq(p\rq_\alpha)$, and at the same time prepares the new pointer $p_\beta$ for the next function $\delta_\beta$ with input $\xvec\rq(p_\beta)\in M^{m_\beta}$.

Note that the two important features of the UMM, namely parallelism and polymorphism, are clearly embedded in the definition of the set of functions $\delta_{\alpha}$. Indeed the UMM, unlike the UTM, can have more than one transition function $\delta_\alpha$ (functional polymorphism), and any function $\delta_\alpha$ {\it simultaneously} acts on a set of memprocessors (intrinsic parallelism). Another difference from the UTM is that the UMM does not distinguish between states of the machine and symbols written on the tape. Rather,  this information is fully encoded in the states of the memprocessors. This is a crucial ingredient in order to build a machine capable of performing computation {\it and} storing of data on the same physical platform.

Another important remark is that unlike a UTM that has only a {\it finite}, discrete number of states and an unlimited amount of tape storage, a UMM can operate, in principle, on an {\it infinite} number of continuous states, even if the number of memprocessors is finite. The reason being that each memprocessor is essentially an analog device with a continuous set of state values\footnote{Of course, the actual implementation of a UMM will limit this continuous range to a discrete set of states whose density depends on the experimental resolution of the writing and reading operations.}.

Finally, it can be noticed that the formal definition of memprocessor~(\ref{memprocdef}) and network of memprocessors~(\ref{memprocnetdef}) is compatible with the function $\delta_\alpha$ defined in \eqref{functUMM} and \eqref{delta}. In fact, the topology and evolution of the network is associated with the stimuli $\svec$, while the control unit defines all possible
$\delta_\alpha\in\Delta$ in the sense that those can be obtained by applying a certain signal $\svec_\alpha$ (which  selects the index vector $p_\alpha$) to the network. The network evolution then determines $\xvec\rq{}$ while $\beta$ and $p_\beta$ (or better $\svec_\beta$) are defined by the control unit for the next processing step.

This more physical description points out a relevant peculiarity of the UMM. In order to implement the function $\delta_\alpha$ (i.e., the computation) the control unit works only on (sends the stimuli $\svec$ to) $\xvec(p_\alpha)$, and the memprocessor network under this stimulus changes the states of $\xvec(p_\alpha\rq{})$ into $\xvec\rq{}(p_\alpha\rq{})$. Therefore, even if $\dim(p_\alpha)=1$, which is the case of the most simple control unit acting on only one memprocessor, we can still have intrinsic parallelism.

\subsection{Proof of universality}

In order to demonstrate the universality of UMMs we provide an implicit proof: we prove that a UMM can simulate any UTM, being this a sufficient condition for universality.

Proving that the UMM can simulate any UTM is actually quite a simple task. We refer to the definition~(\ref{UTMdef}) of UTM given in section \ref{UTM}. First, we take a UMM having the memprocessor states defined by $M=Q\cup\Gamma$. One memory cell is located by the pointer $j_s$ while the (infinite) remaining cells are located by the pointer $j=...,-k,...,-1,0,1,...,k,...$. We further define the array of pointers $p$ (unique element of $\cal P$) defined by $p=\{j_s,j\}$. We use the cell $j_s$ to encode the state $q\in Q$, and in the other cells we encode the symbols in $\Gamma$.

We now take $\Delta$ to be composed by only one function $\bar\delta[\xvec(p)]=(\xvec\rq(p),p\rq)$, where we have suppressed the output index $\beta$ because the function is only one. The new states $\xvec\rq$ written by $\bar\delta$ are written according to the table of the UTM (i.e., according to $\delta$ of Eq.~(\ref{deltaUTM})), and in particular in $\xvec\rq(j_s)$ we find the new state the UTM would have, and in $\xvec\rq(j)$ we find the symbol the UTM would write on the tape. Finally, the new pointer $p\rq$ is given by $p\rq=\{j_s,j\rq\}$ where $j\rq=j$ if $\delta$ of UTM requires No shift, $j\rq=j+1$ if Right shift, and $j\rq=j-1$ if Left shift. Finally, following this scheme, writing on $\xvec(j_s)$ the initial state $q_0$ and the initial symbols $\Sigma$ where required, the UMM with $\Delta=\bar\delta$ simulates the UTM with $\delta$.

We have thus shown that a UMM is {\it Turing-complete}, namely it can simulate {\it any} Turing machine (whether deterministic or not). Note, however, that the reverse is not necessarily true. Namely, we have not demonstrated that a UTM can simulate a UMM, or, equivalently, we have not demonstrated that a UMM is Turing-equivalent. It is worth pointing out that, if we could prove that a UMM is not Turing-equivalent, some (Turing) undecidable problems, such as the halting problem \cite{turing_book}, may find solution within our UMM paradigm, thus contradicting the Church-Turing hypothesis~\cite{turing_book}. Although this is an intriguing--albeit unlikely-- possibility, we leave its study for future work.

\section{Properties of UMMs}\label{UMMSecP}

\subsection{On the parallel computing of UMMs}

\begin{figure}
\centerline{
\includegraphics[width=\columnwidth]{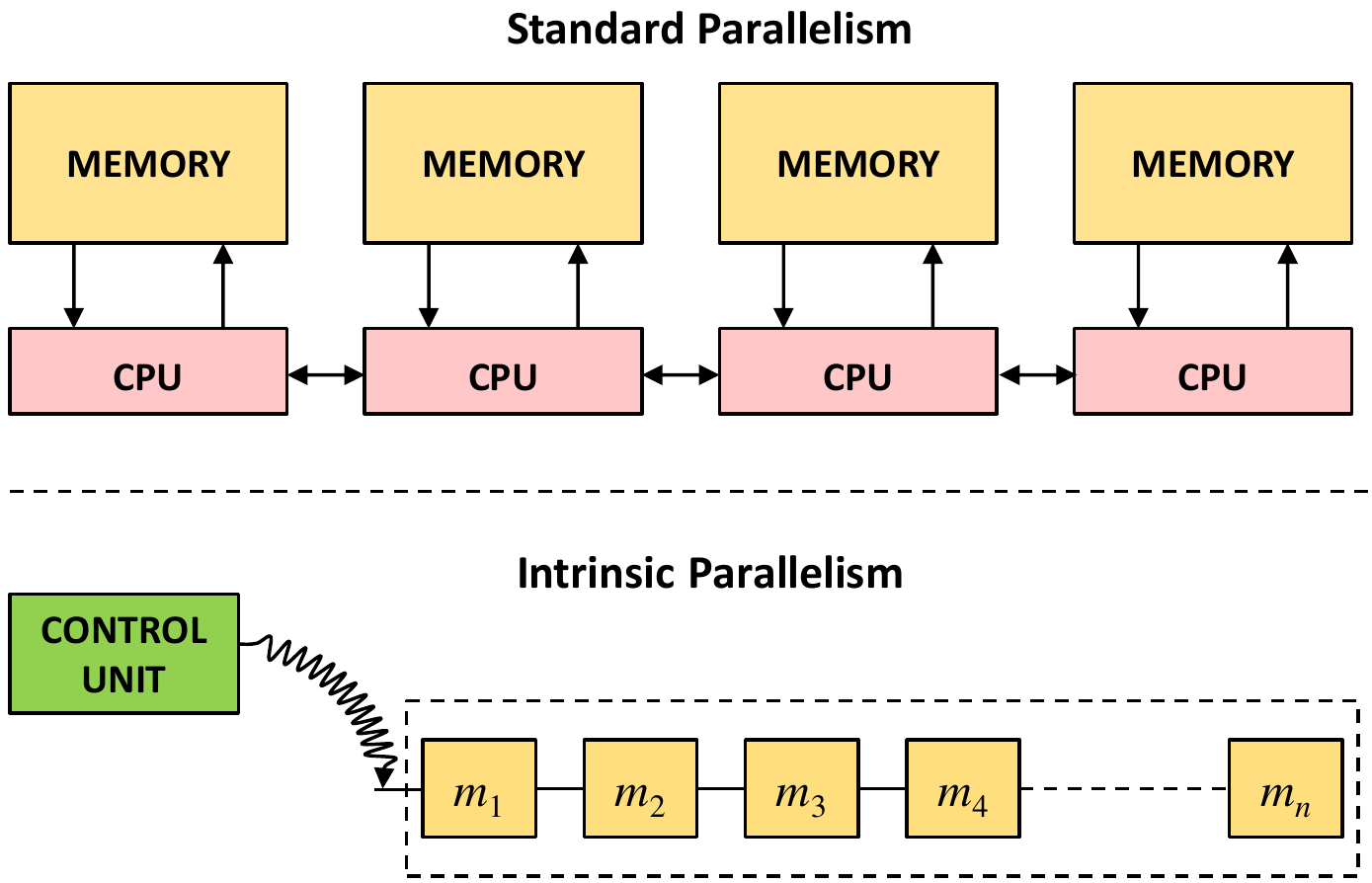}}
\caption{\label{parallelism}Sketch of comparison between standard parallel architectures and intrinsic parallelism of UMMs.}
\end{figure}

Parallelism in computing is an important issue from both a theoretical and a practical point of view. While the latter is more important for algorithm implementation in actual parallel architectures---multi-core CPU, CPU or GPU clusters, vector processors, etc. (see top panel of Figure~\ref{parallelism} for a sketch)---the former approach is essential in complexity theory. Roughly speaking, practical parallel computing involves a fixed number of processing units working in parallel, or a number growing at most polynomially with respect to the dimension of the input. This means that the NP problems still have a NP solution when implemented in practical parallel architectures.

On the other hand, the UMM approach to parallel computing can account for a number of processing units working in parallel that can grow also exponentially, if required. In order to understand this concept better we first briefly discuss two important and representative types of parallel machines before discussing the parallelism characterizing the UMM.

The first machine we consider is the Non-Deterministic Turing Machine (NTM) \cite{complexity_bible}. An NTM differs from a (deterministic)
 Turing machine (TM) for the fact that, in general, the transition function $\delta$ does not have a uniquely specified three-tuple $(q\rq,\gamma\rq,s\rq)\in Q\times\Gamma\times\{ L,N,R\}$ for each $(q,\gamma)\in Q\backslash F\times\Gamma$ \cite{complexity_bible}. Therefore, unlike the TM, the NTM needs to be coupled with a way to choose among the possible actions of $\delta$. This can be done in two different (equivalent) ways: either there is an \textit{oracle} that picks the transition that eventually leads to an accepting state---if there is such a transition---or the machine branches into many paths each of them following one of the possible transitions \cite{complexity_bible}. The second interpretation of a NTM is that of a parallel machine with an exponentially growing number of processing units at each level of the solution tree~\cite{complexity_bible}. Clearly, the NTM is only an ideal model, impossible to reproduce in practice. Nevertheless, it is of great importance because the most famous unsolved question in Computer Science, the NP=P problem, is directly related to the question
 of whether the (deterministic) TM can simulate in polynomial time a NTM.

The second machine that boasts parallelism is the Quantum Turing machine (QTM) \cite{QTM}. The QTM is an ideal machine used to model the behavior of a possible Quantum Computer (QC). One of the essential differences from the classical TM is that a QTM operates on a superposition of quantum states, each of them encoding a certain amount of data. For example, in the most famous algorithm for quantum computation, the Shor\rq{}s algorithm for integer factorization \cite{Shor_0,Shor_1}, the QC employs a registry made of $n$ qubits encoding a superposition of $2^n$ states. In this way, using quantum gates, a QC can process at the same time $2^n$ states by using only $n$ qubits. Therefore, a QTM is a massively-parallel machine that can process at the same time an exponential number of states using a memory that grows linearly. However, even
with its massive parallelism, it is not yet proven that a QTM is equivalent to an NTM \cite{QI_bible}. (Note that claims to the contrary have been disproved \cite{QI_bible}.) On the other hand, it is conjectured but not proven that the NTM is a much more powerful machine than a QTM \cite{QI_bible}. For example, NTMs can solve NP--complete problems in polynomial time, but there is no evidence that QTMs could accomplish the same task.

Now, let us discuss the parallelism characterizing the UMM and the computation power arising from it. Following the sketch in the bottom panel of Figure~\ref{parallelism} and Figure~\ref{solution_tree}, the intrinsic parallelism characterizing UMMs is the consequence of the physical coupling between memprocessors. When the control unit sends an input signal to one or a group of memprocessors, the coupling induces dynamics to a larger group of memprocessors.  Therefore the computation involves not just the memprocessors directly connected to the control unit, but up to the entire network. In principle, with only one input signal we can then induce massively-parallel computation in the entire network as also discussed in Sec.~\ref{formalUMM}.

This is not the only consequence of the physical coupling among memprocessors. As we discuss in the Sec.~\ref{information_overhead}, using the coupling between memprocessors, we can encode information not necessarily proportional to the number of memprocessors, but up to an exponential amount in terms of the number of memprocessors. Therefore, by using the intrinsic parallelism, we can also manipulate the information encoded in the memprocessor network {\it all at once}. As we prove in the next sections, the UMMs -- thanks to their intrinsic parallelism -- can solve NP-complete problems in polynomial time using an exponential number of memprocessors or, boosted by the information overhead, using only a polynomial number of memprocessors.

\subsection{UMMs solve NP--complete problems in polynomial time}

We consider an NP--complete problem whose known fastest solving algorithm requires a solution tree {\it exponentially} growing with the dimension $n$ of the input. Such algorithm, implemented in a deterministic TM would require exponential time (steps) with respect to $n$ in order to be solved.

We consider a formal description of the solution tree assuming that the branch of the tree leading to the solution takes a polynomial number of iterations with respect to $n$, i.e., $N_s=P(n)$ being $N_s$ the number of iterations and $P(n)$ a polynomial function. At each iteration $i$ we can assume that each branch splits into $M_i$ new branches, so the complete solution tree has a total number of nodes $N_{nodes}\leq\sum_{k=1}^{N_s}\prod_{i=1}^{k}M_i$ (where the situation ``$<$'' arises when some branches end with no solution before the $P(n)$--th iteration).

\begin{figure}
\centerline{
\includegraphics[width=1\columnwidth]{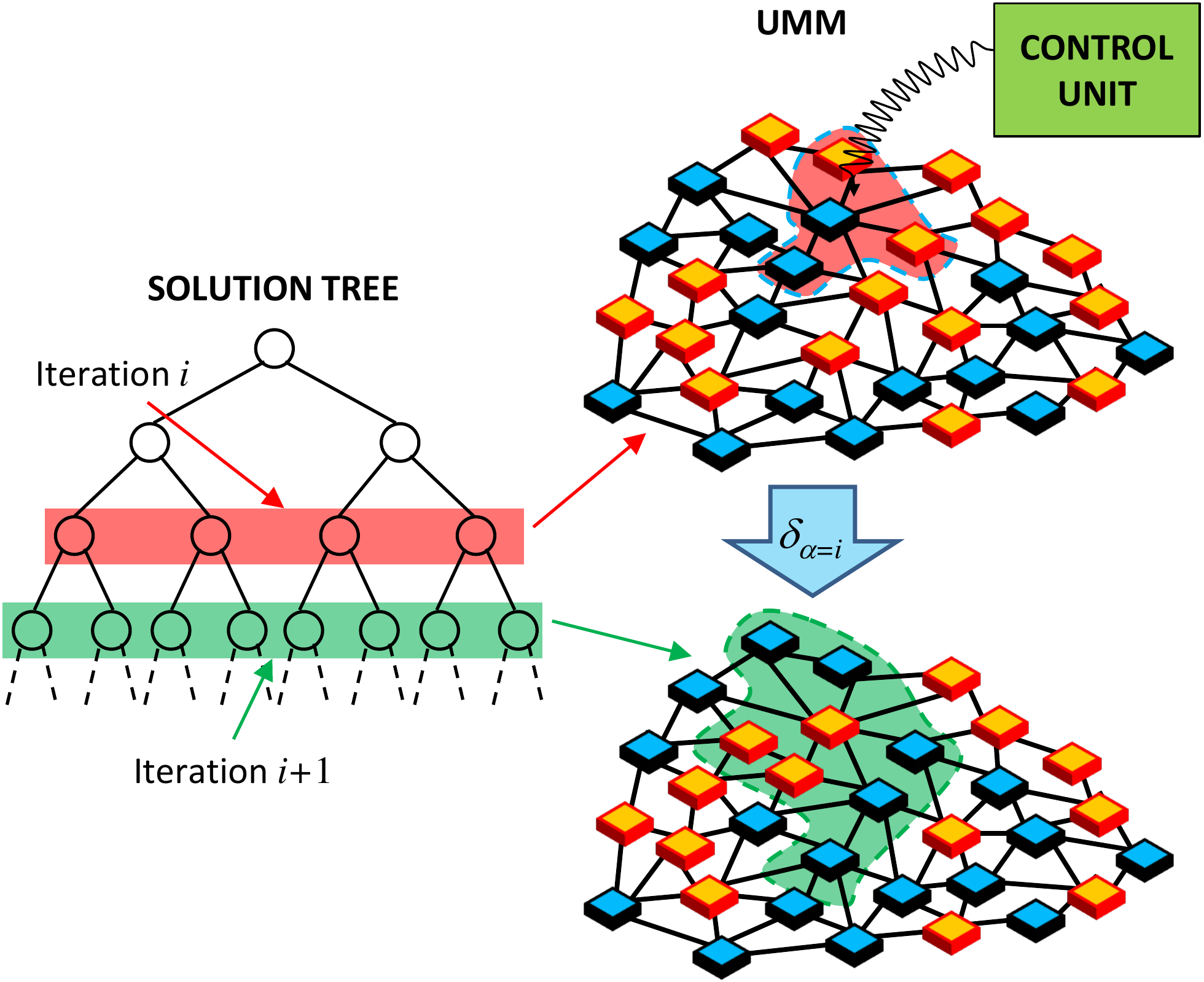}}
\caption{\label{solution_tree}Left panel: a typical solution tree for an NP--complete problem. Right panel: Sketch of the UMM transition function to solve it. In this particular figure, the computational memory is formed by a simple network of 2-state memprocessors. The control unit sends the input signal to a group of memprocessors. The signal induces computation into the network and for each input signal sent by the control unit the UMM computes an entire iteration of the tree.}
\end{figure}
\begin{figure*}
\centerline{
\includegraphics[width=1.9\columnwidth]{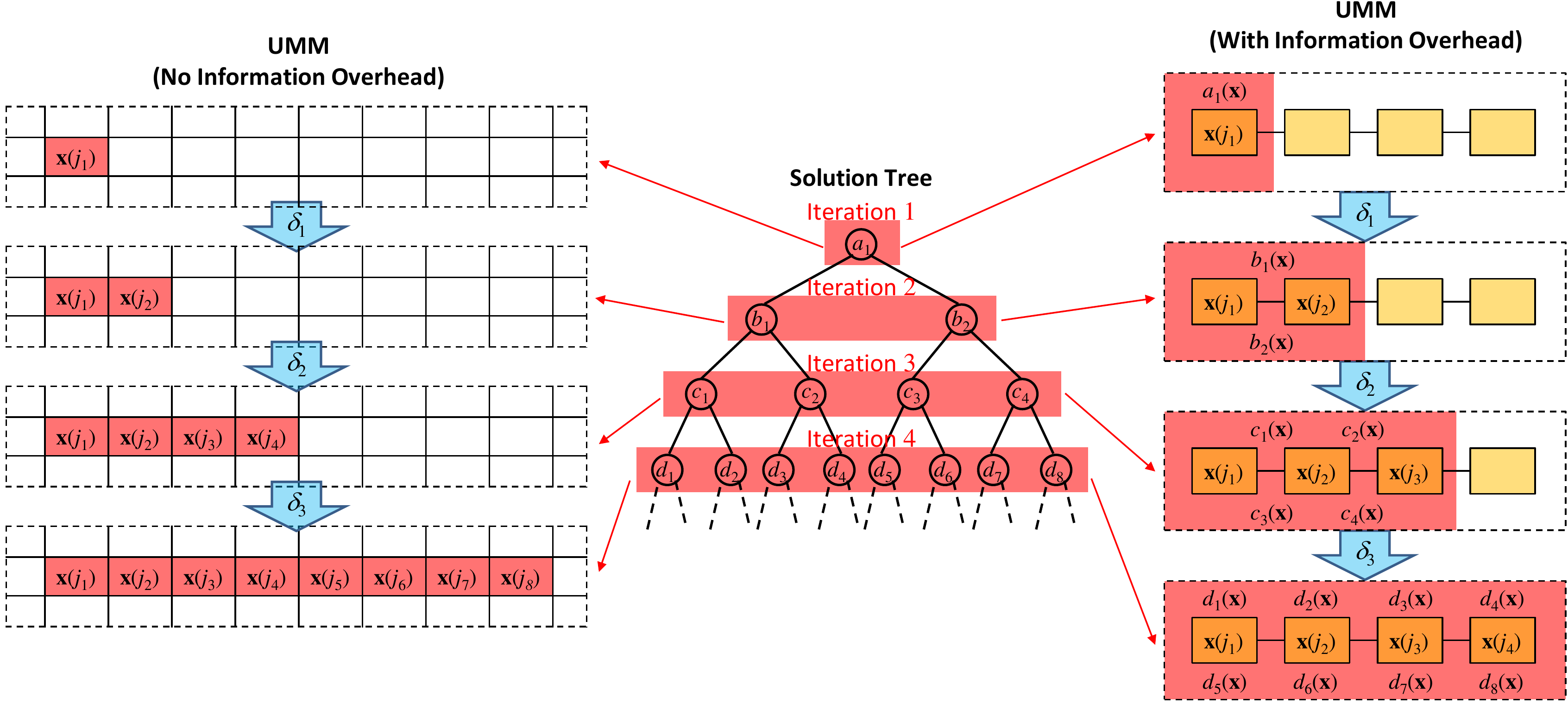}}
\caption{\label{solution_tree2}Solution tree of a NP--complete problem implemented in a UMM. The transition from the iteration $i$ to the iteration $i+1$ of the solution tree is computed by the function $\delta_\alpha=\delta_i$ acting on the group of memprocessors encoding the iteration $i$ to give the iteration $i+1$ all at once (intrinsic parallelism). UMMs that do not employ the information overhead encode the information of each node of the solution tree into a proportional number of memprocessors (in left panel each node is encoded into a memprocessor), while UMMs employing information overhead can encode the information of a group of nodes (for example an entire iteration) in the collective state of a group of memprocessors (right panel).}
\end{figure*}
Following this picture, if the simplest tree has an average number $\bar M$ of branch splits, then the time required by a TM to explore the solution tree, that is proportional to the number of nodes $N_{nodes}$, is of the order of $\bar M^{N_s}=\bar M^{P(n)}$, i.e., exponentially growing with respect to the dimension $n$ of the input. On the contrary, we now prove that the UMM we have introduced in Sec.~\ref{formalUMM} can find the solution within a time proportional to $P(n)$.

In order to prove this claim, we consider the UMM that operates following the scheme of figure~\ref{solution_tree}. The control unit sends a signal input to a group of interconnected memprocessors encoding the iteration $i$ of the solution tree to compute the iteration $i+1$ all at once. Thus the entire computation process can be formalized following the scheme of Figure~\ref{solution_tree2}. We consider a set $\Delta$ of functions $\delta_\alpha$ with $\alpha=1,...,P(n)$ such that $\delta_\alpha[\xvec(p_\alpha)]=(\xvec\rq(p_{\alpha+1}),\alpha+1,p_{\alpha+1})$ where $\xvec(p_\alpha)$ encodes all the nodes belonging to the iteration $\alpha=i$ of the solution tree into at most $\dim(p_\alpha)\leq\prod_{i=1}^{\alpha}M_i$ memprocessors, and $\xvec\rq(p_{\alpha+1})$ encodes all the nodes belonging to the iteration $\alpha+1=i+1$ of the solution tree into at most $\dim(p_{\alpha+1})\leq\prod_{i=1}^{\alpha+1}M_i$ memprocessors. It is also worth noticing that in the UMM case, the situation ``$<$'' does not arise only when some branches end with no solution before the $P(n)$--th iteration. Rather, by  exploiting the information overhead we discuss later and sketched in the right panel of Figure~\ref{solution_tree2}, the data encoded into $N_{nodes_\beta}\leq\prod_{i=1}^{\beta}M_i $ can be encoded in a number of memprocessors $\dim(p_{\beta})\leq N_{nodes_\beta}$ (see section~\ref{information_overhead}).

Since $\delta_\alpha$ corresponds to only {\it one computation step} for the UMM, it will take a time proportional to $P(n)$ to find the solution. Therefore, in principle, the UMM can solve an NP--complete problem in a time proportional to $P(n)$, i.e., {\it the UMM has the same computational power of a NTM}.

Using this simplified scheme for the solution of an NP--complete problem, we can highlight the fundamental difference between the UMM and the NTM. The NTM (in the ``exploring-all-possible-paths'' picture) needs, in principle, {\it both} a number of processing units {\it and} a number of tapes exponentially growing in order to explore all possible solution paths~\cite{complexity_bible}. On the contrary, a UMM, in order to find a solution of an NP--complete problem in polynomial time, needs only a number of memprocessors at most exponentially growing, since it can process an entire iteration of the solution tree in just one step. However, there is yet the final property of UMMs, the information overhead, that allows us to reduce this exponentially growing number substantially, even to a polynomial order, making the practical implementation of a UMM even more appealing. Note also that the above proof is valid for {\it any} problem whose solution can be cast into an exponentially growing tree, whether the problem is NP--complete or NP--hard. In Sec.~\ref{SSPSec} we indeed provide the polynomial solution of the sub-set sum problem in both its decision and optimization versions.

\section{Information Overhead}\label{information_overhead}

As already anticipated, the intrinsic parallelism and functional polymorphism we have discussed above are not the only features that resemble those employed by a typical brain. Indeed, a UMM boasts an additional property we name \textit{information overhead}. In order to explain this third peculiar feature, we first recall how the information is stored in modern computers. In these devices, the information is encoded in bits, which ultimately correspond to the threshold state of some electronic device (patterns of magnetization in a magnetizable material, charge in capacitors, charge in a Floating-Gate MOSFET, etc.) \cite{computer_architecture_book}. Irrespective, the memory cells in modern computers are always understood as independent, non-interacting entities. Therefore, the quantity of information that can be stored is defined from the outset and can not exceed an amount proportional to the number of employed memory cells.

On the other hand, if the memory cells are connected or can interact in some ways, this limitation does not hold. In fact,
although the issue is still under debate, it is thought that the brain stores information by changing the way in which the neurons communicate through their synapses \cite{hebb_book}. Therefore, the storing of information is not a local action involving a single neuron or synapse. Rather it is a {\it collective} behaviour of a group of interacting neurons \cite{kohonen_book}.

Following this picture, the brain information storage capacity cannot be simply proportional to the number of neurons or synapses: a more complex relation must exist. For example, a person affected by hyperthymesia is not able to erase his/her memory \cite{06_parker,12_leport}, thus demonstrating the capability of our brain to store such a large quantity of information whose limits, currently, cannot even be estimated. In other words, it is reasonable to assume that our brain is capable of some sort of {\it information overhead}: it can store a quantity of information not simply proportional to the number of neurons or synapses.

The UMMs we have defined satisfy a similar property. In order to clarify this issue, we describe two examples of information overhead arising in interconnected memprocessors easily reproducible with memelements, e.g., memristors or any other memory cell. Here, we do not specify the actual physical coupling between the memprocessors, just note that it is some physically plausible interaction. For instance, this interaction could be simply provided by the direct local wiring between the memprocessors as we do in Sec.~\ref{SSPSec}, which indirectly gives rise to the non-local collective dynamics as determined by Kirkshoff's laws.

\subsection{Quadratic Information Overhead}

Formally, the information stored in a message $m$ can be quantified by the Shannon\rq{}s self-information $I(m)=-\log_2p(m)$ where $p(m)$ is the probability that a message $m$ is chosen from all possible choices in the message space $M$ \cite{information_book}. The base of the logarithm only affects a scaling factor and, consequently, the units in which the measured information content is expressed.
 For instance, the logarithm base 2 measures the information in units of bits.

 Now, let us consider the message space composed by messages $m$ formulated in the following way: given the set $G$ composed of $n$ different integer numbers with sign and considering all the possible sums of $2$, $3$, ..., $k$ of them taken only once and $k\leq n$. We then define each message $m$ as that containing both the numbers that we are summing and their sum. Therefore, the message space $M$ is composed of $\sum_{j=2}^{k}\tbinom{n}{j}$ equally-probable messages $m$ with Shannon\rq{}s self-information $I(m)=\log_2\sum_{j=2}^{k}\tbinom{n}{j}$.

\begin{figure}
\centerline{
\includegraphics[width=0.8\columnwidth]{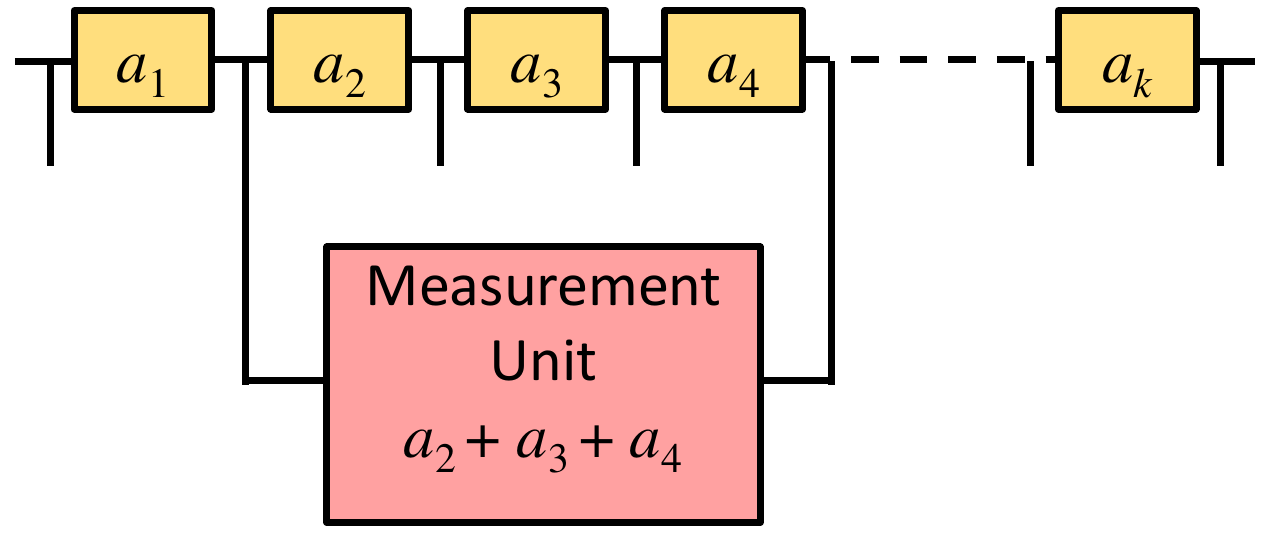}}
\caption{\label{connected_cells}System of $k$ connected memprocessors in series. By using intermediate connections it is possible to directly measure the sum of intermediate cells.}
\end{figure}

  Let us now consider $k$ memprocessors capable of storing any integer number belonging to $M$ and we consider two different scenarios: {\it i)} $k$ independent memprocessors acting as simple memory cells, and {\it ii)} $k$ memprocessors connected in series as in figure \ref{connected_cells}. In the first scenario, by defining $\lfloor{x}\rfloor$ the floor function operator that rounds $x$ to the nearest integer towards 0, we can store at most $\lfloor{k/3}\rfloor$ messages $m$, i.e., the messages $m$ containing two numbers (i.e., the smallest number of numbers) and their sum.

On the contrary, in the second scenario, by storing $k$ different numbers from the $n$ numbers, we can also directly read their sums as in figure \ref{connected_cells}. Therefore, in this case we can store $n(n-1)/2$ messages $m$, namely a {\it quadratic} (in the memprocessors) number of messages.

From this simple example we can already highlight several points. First of all, we have shown that by using interconnected cells we can store (compress) a quadratic amount of information with respect to the number of memprocessors employed. This information overhead is essentially a {\it data compression
 that is already present directly in memory}. In fact, we can store the same amount of information in the $k$ independent memprocessors provided we couple them with a CPU performing sums, i.e., standard data compression requiring memory plus a program implemented in a CPU to decompress data.

 We can however easily implement in hardware the interconnected memprocessors using, e.g., memristor crossbars \cite{08_strukov}, each memristor storing a number in $G$ that is linearly encoded through the resistance of the memristor. In order to read the sum we can simply apply a small voltage to the interconnected memprocessors and measure the current which is proportional to the inverse of the sum of the resistances involved. It is worth noticing that this practical implementation of information overhead can already be used to speed up typical NP-complete problems such as the subset-sum problem (SSP) we discuss below in Sec.~\ref{SSP}. However, with this strategy the number of memprocessors can only be reduced polynomially. Although this is not a formal problem, it is definitely a practical limitation.

\subsection{Exponential Information Overhead}\label{exp_IO_sec}

\begin{figure}
\centerline{
\includegraphics[width=1\columnwidth]{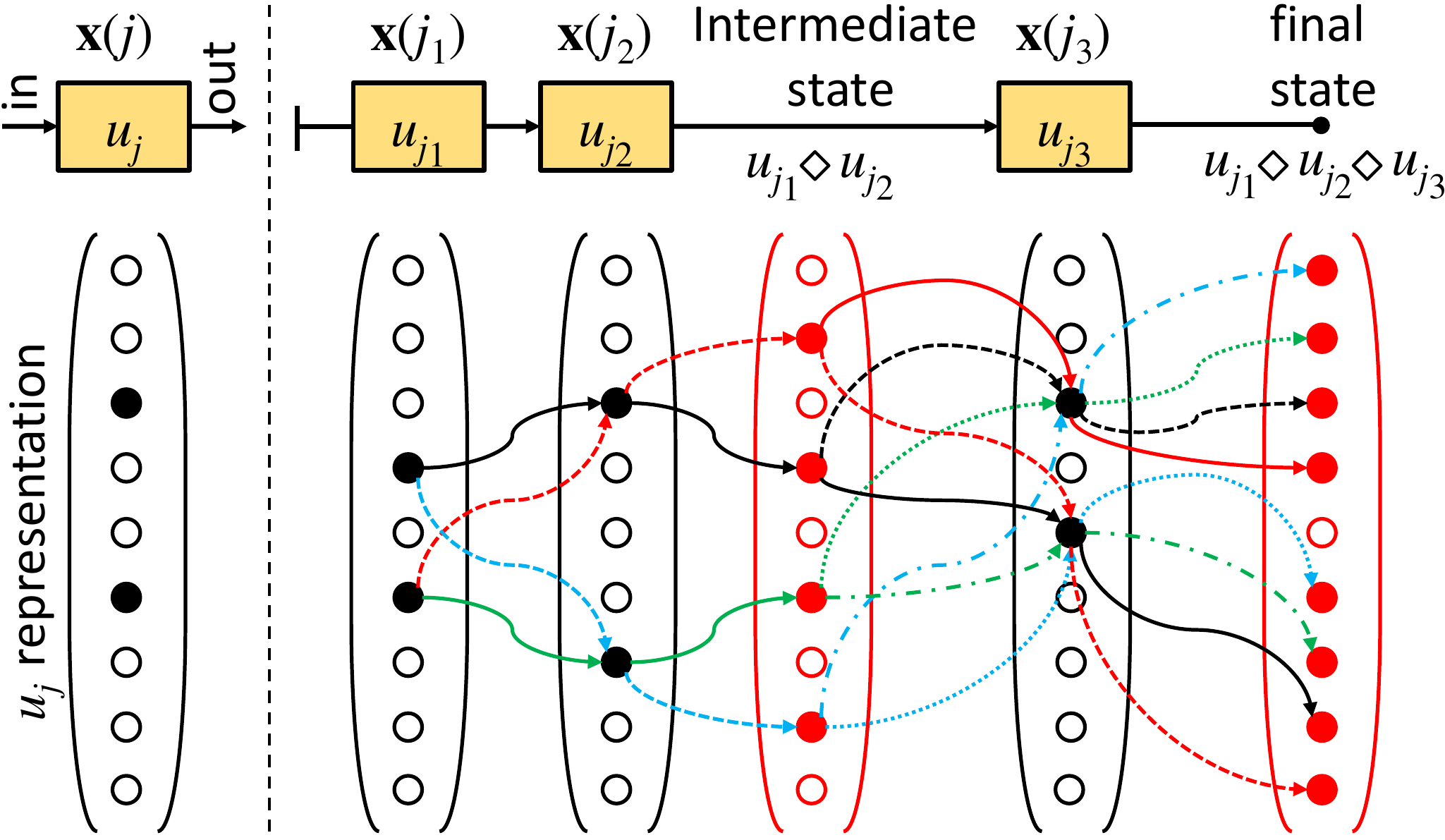}}
\caption{\label{exp_IO}Left panel: two-terminal memprocessor and representation of its internal state. Right panel: sketch of three interconnected memprocessors.}
\end{figure}

Another example of information overhead that we present allows instead an exponential reduction of memprocessors. To illustrate what we mean by this, let us consider a set of $n$ memprocessors defined by $\xvec(p)=(\xvec(j_1),\dots,\xvec(j_n))$. Each memprocessor state $\xvec(j)=u_j$ can store data encoded into its internal state vector $u_j\in M_a$. We describe the internal state $u_j$ as a vector with (possibly infinite) components $\{u_j\}_h$  and we assume the memprocessor as a two terminal \lq\lq{}polarized\rq\rq{} device in the sense that the right terminal and the left terminal, when connected to other memprocessors, physically interact in different ways, thus we label as \lq\lq{}in\rq\rq{} and \lq\lq{}out\rq\rq{} the two different terminals (see left panel of figure \ref{exp_IO}). We assume the blank state of the memprocessors corresponds to all components of $u_j$ equal to 0 (empty circles of figure \ref{exp_IO}). In order to store one datum we set one of the components of $u_j$ different from 0 (full circles of figure \ref{exp_IO}), and any memprocessor can simultaneously store a finite number of data. Moreover, we assume that there exists a device that, once connected to a memprocessor, can read each single component of $u_j$.

The former description of the memprocessor provides the picture of a single non-interacting memprocessor. However, we complete this picture by including a physical interaction scheme between different memprocessors. We consider two memprocessors connected through the \lq\lq{}out\rq\rq{} and \lq\lq{}in\rq\rq{} terminals of the first and second memprocessor, respectively.
In addition, we assume that the device that reads the state of a single memprocessor can be used to read the global state of the two connected memprocessors as well. Therefore, the device reads the global state of two connected memprocessors that we describe as $u_{j_1,j_2}=u_{j_1}\diamond u_{j_2}$ where $\diamond:\mathbb{R}^d\times\mathbb{R}^d\rightarrow\mathbb{R}^d$ is a commutative and associative operation with $d=\dim(u_j)$ and it is defined through
\begin{equation}
\{u_{j_1}\diamond u_{j_2}\}_{h\star k}=\{u_{j_1}\}_h\ast\{u_{j_2}\}_k
\end{equation}
where $\star:\mathbb{Z}\times\mathbb{Z}\rightarrow\mathbb{Z}$ and $\ast:\mathbb{R}\times\mathbb{R}\rightarrow\mathbb{R}$ are two commutative and associative operations with properties $h\star 0=h$ and $x\ast 0=0$. Although one can envision
non-commutative and non-associative types of physical interactions, for the
sake of simplicity we restrict our analysis only to commutative and associative operations, leaving the more complex ones
for future studies.

In addition, if there are two couples of indexes $h,k$ and $h\rq{},k\rq{}$ such that $h\star k=h\rq{}\star k\rq{}$, then $\{u_{j_1,j_2}\}_{h\star k}=\{u_{j_1}\diamond u_{j_2}\}_{h\star k}\oplus\{u_{j_1}\diamond u_{j_2}\}_{h\rq{}\star k\rq{}}$ where $\oplus:\mathbb{R}\times\mathbb{R}\rightarrow\mathbb{R}$ is another commutative, associative operation such that $x\oplus 0=x$. Since all the defined operations are commutative and associative, we can straightforwardly iterate the procedure to any number of connected cells (see right panel of figure \ref{exp_IO} for a sketch of three interacting memprocessors).

Given a set $G=\{a_{1},\dots,a_{n}\}$ of integers we now define a message $m=\left(a_{\sigma_1}\star\dots \star a_{\sigma_k}\right)\cup\{a_{\sigma_1},\dots , a_{\sigma_k}\}$ where $\{\sigma_1,\dots , \sigma_k\}$ is a set of indexes taken from all possible different subsets of $\{1,\dots,n\}$ so that the message space $M$ is composed of $\sum_{j=0}^{n}\tbinom{n}{j}=2^n$ equally-probable messages $m$ with Shannon\rq{}s self-information $I(m)=\log_2(2^n)=n$. Now, taking $n$ memprocessors, for each one of them we set only the components $u_{j_0}$ and $u_{j_{a_h}}$ different from zero with $h\in \{1,\dots,n\}$. In this way, in each memprocessor one element of $G$ is encoded. On the other hand, by following the picture of physically interconnected memprocessors described previously, if we read the global state of the interconnected cells we find all possible messages $m$, which means that $n$ interconnected memprocessors can encode (data compress) $2^n$ messages simultaneously.

In section \ref{SSPSec_exp} we discuss how to use in practice this exponential information overhead to solve the subset-sum problem in polynomial time using only $n$ memprocessors. Here, we stress again that, even if we have used operations to define both the quadratic and the exponential overhead, these operations represent the {\it physical coupling} between the different memprocessors and {\it not} the actual computation defined by the functions $\delta_\alpha$ (Eq.~(\ref{functUMM})) of a UMM. In fact, while the computation changes the state of some (or all) memprocessors, the information overhead -- which is related to the {\it global} state of physically interacting memprocessors -- does not.

\section{The subset-sum problem}\label{SSPSec}

In order to exemplify the computational power of a UMM, we now provide two actual UMM algorithms to solve the NP--complete \textit{subset-sum problem} (SSP), which is arguably one of the most important problems in complexity theory \cite{complexity_bible}.
It can be defined as follows: if we consider a finite set $G\subset\mathbb{Z}$ of cardinality $n$, we want to know whether there is a non-empty subset $K\subseteq G$ whose sum is a given number $s$.

\subsection{DCRAM-based algorithm}

\begin{figure}
\centering
\subfigure
{\includegraphics[width=1\columnwidth]{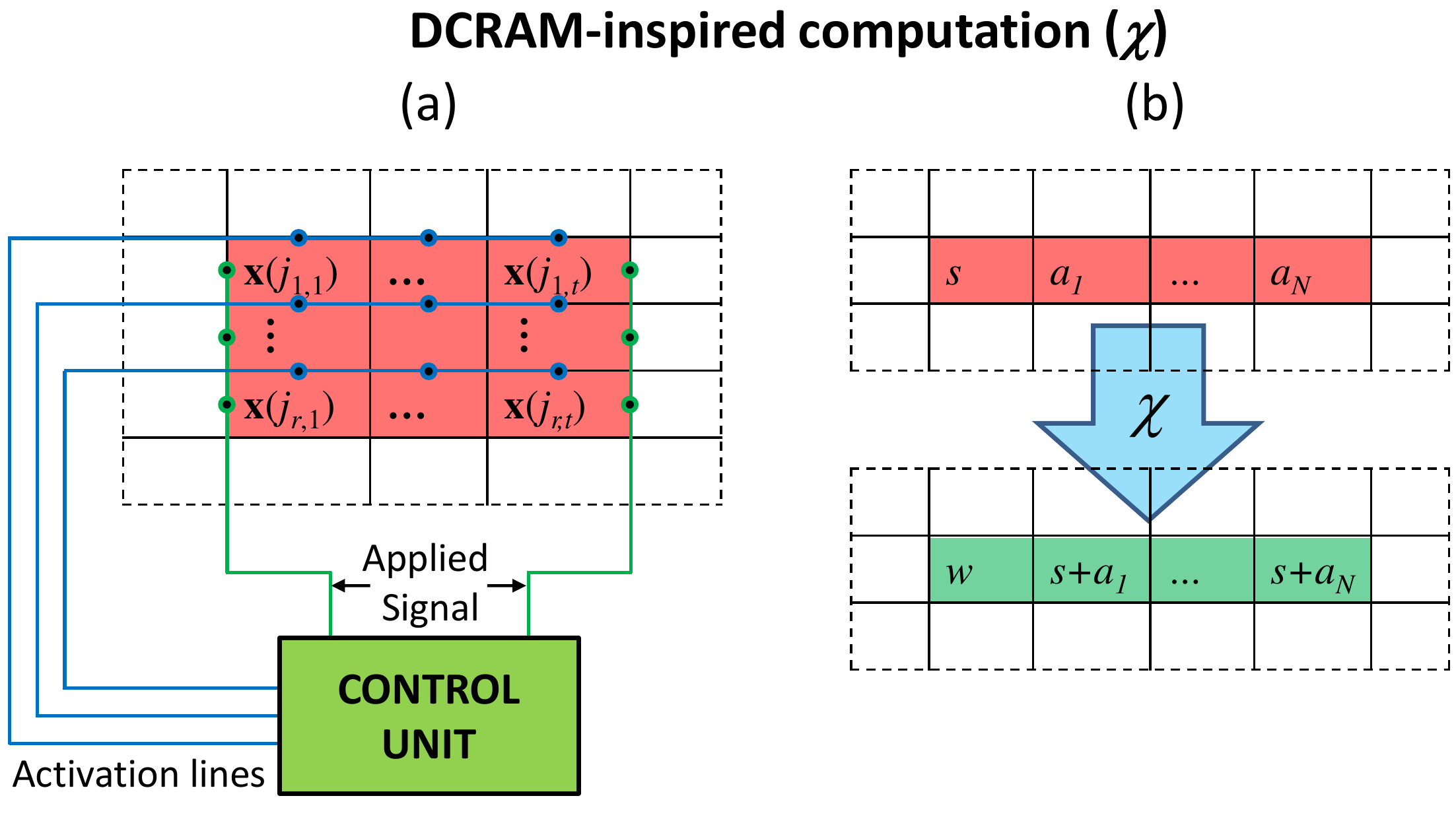}}
\subfigure
{\includegraphics[width=1\columnwidth]{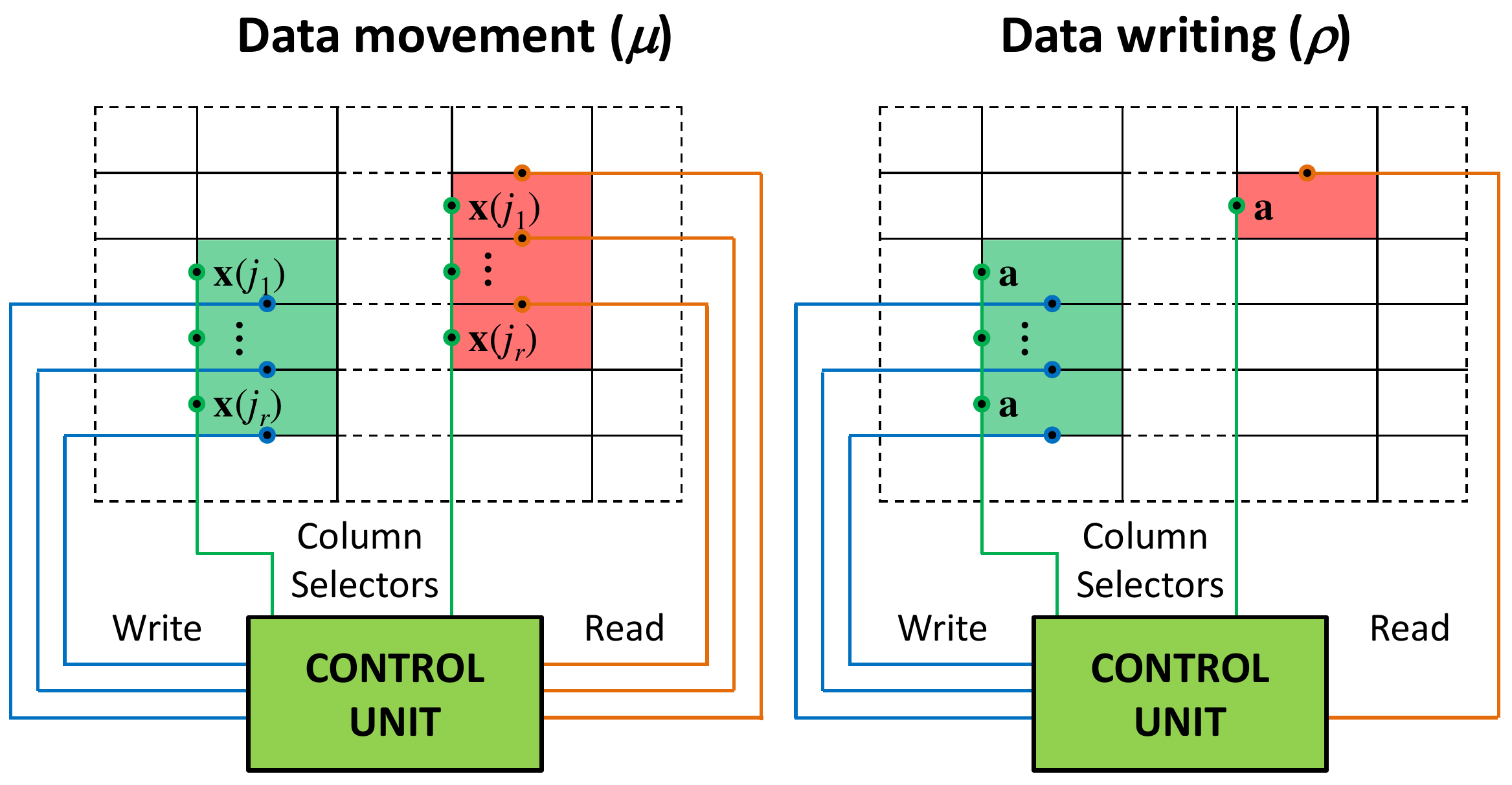}}
\caption{\label{operations}The three possible operations that can be performed by the DCRAM-inspired machine. The $a_j$,  $s$, $w$ and $\mathbf a$ are integers with sign encoded in the states $\xvec$.}
\end{figure}

In order to show the first practical algorithm implementation that can solve in polynomial time the SSP using a UMM, we consider a possible realization of a UMM through an ideal machine. This machine is inspired by the Dynamic Computing Random Access Memory (DCRAM) introduced by the authors in Ref. \cite{DCRAM}. In the present work, we employ an idealization of a DCRAM in the sense that we do not specify the actual electronic devices capable of reproducing exactly the machine we are going to present. Nevertheless, this does not exclude the possibility that such a machine can be realized in practice. Moreover, while the DCRAM designed in \cite{DCRAM} is digital, i.e., it has the state of a single memprocessor belonging to $M=M_d$, the current DCRAM has memprocessor states belonging to $M=M_a$ in order to perform the operation $\chi$ described in Fig.~\ref{operations}--(a). It is worth noticing that, even if $\chi$ is the analog computation between integers, it can be simply realized with boolean gates, so, in principle, even with digital DCRAM-inspired machines.

The DCRAM-inspired machine has memprocessors organized in a (possibly infinite) matrix architecture. Through a control unit we can perform three possible operations (see figure~\ref{operations}).
\begin{itemize}
\item[$\chi$:] This is the actual computation. By using the activation lines (similar to word lines of standard DRAM) to select the rows to be computed, and applying a signal at two selected columns (see Fig.~\ref{operations}--(a)), the memprocessors belonging to the same row change their states according to the operation in Fig.~\ref{operations}--(b), i.e., the datum stored in the first column is summed to the data stored in the other columns. Therefore, through this operation, by just applying a single signal, we can compute in a massively-parallel way the data stored in a submatrix of the DCRAM-inspired machine. It is worth noticing that during the computation no data moves, and the control unit does not have any information of the data stored in the memory.
\item[$\mu$:] This is the movement of data. The control unit selects two columns and through read and write lines the data flow from one column to the other. Also in this case, the control unit does not necessarily read and write the data, but it actually connects the cells allowing for data flow.
\item[$\rho$:] This operation is similar to $\mu$ but it replicates one data in a column.
\end{itemize}

\begin{figure}
\centerline{
\includegraphics[width=1\columnwidth]{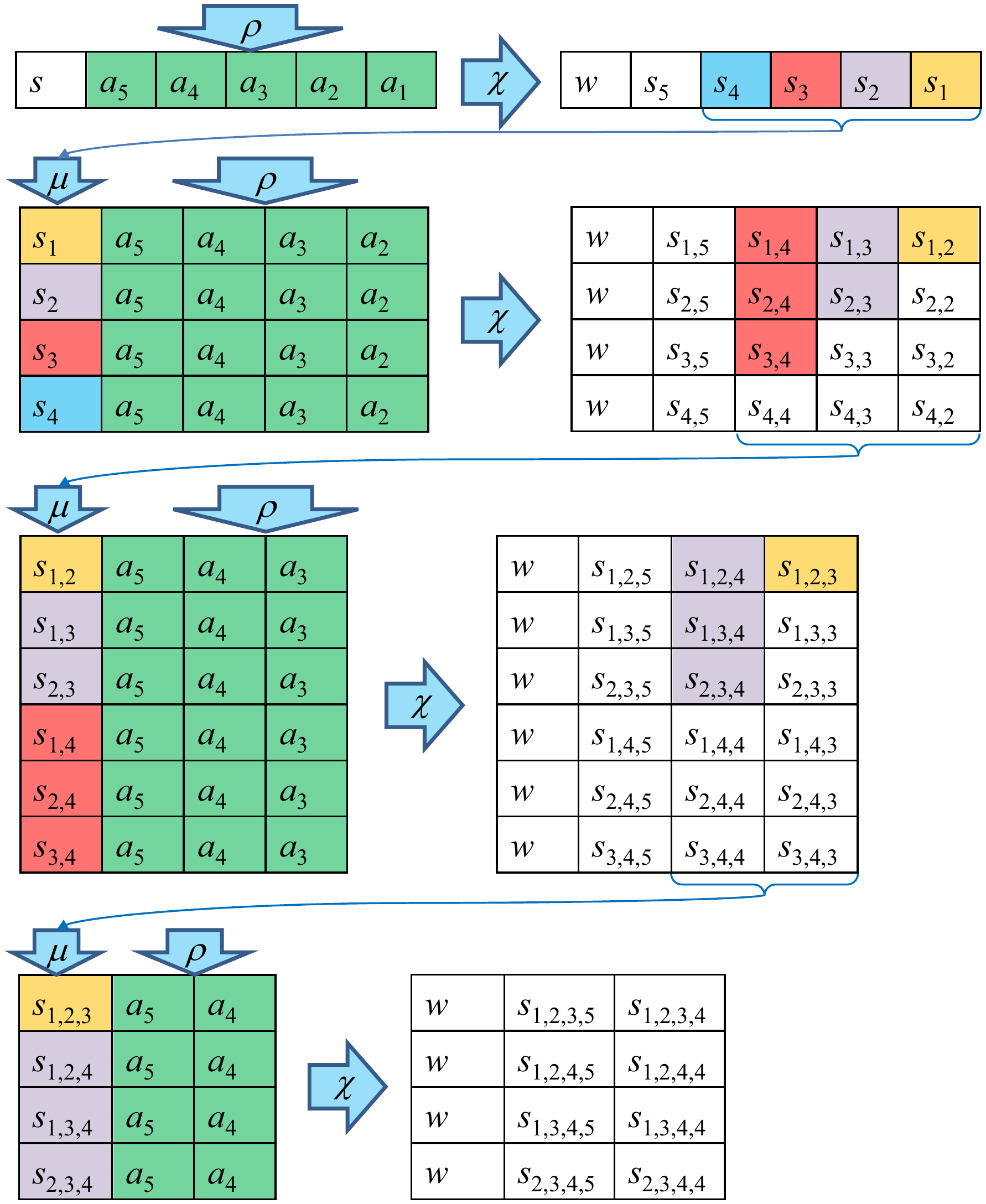}}
\caption{\label{SSP}Solution of the subset-sum problem for a set $G\subset\mathbb{Z}$ composed by $n=5$ numbers. The symbols are: $a_j\in G$, $s=-\sum_{j=1}^na_j$, $s_{j,k}=s+a_j+a_k$ and $w$ are unassigned data. At each computation step, the functions $\rho$ and $\mu$ manipulate in parallel all memprocessors with the same color and the function $\chi$ uses in parallel all the memprocessors of the entire subnetwork reported at each step.}
\end{figure}

Combining these three functions we obtain the transition function $\delta=\mu\circ\chi\circ\rho$ of the machine and we can solve the SSP using only $n-1$ iterations of $\delta$. In order to show how it works, in Fig.~\ref{SSP} the algorithm implementation for a set $G$ of cardinality 5 is reported. At the first step the machine computes the sums of all the possible subsets with cardinality $n-1$. Here, if none is the given number $s$, the procedure continues and the sums of all the possible subsets with cardinality $n-2$ are calculated. The procedure ends if it finds some subsets summing to $s$, or if it reaches the $n-1$--th iteration.

From this scheme it is simple to evaluate the total number of memprocessors needed at each iteration. In fact, considering the iteration $k$, we need $\tbinom{n-1}{k-1}(n+2-k)$ memprocessors. Using the Stirling formula and observing that the binomial $\tbinom{n}{k}$ has maximum in $k=n/2$ (for odd $n$, $k=(n\pm1)/2$), the maximum number of memory cells required by the algorithm is about $(n/2\pi)^{1/2}2^{n-1}$ showing the exponential growth of the number of memory cells.

\subsection{Exponential information overhead-based algorithm}\label{SSPSec_exp}

Another algorithm to solve the SSP can be obtained by exploiting the exponential information overhead discussed in section \ref{exp_IO_sec}. Here, we discuss both its numerical implementation that provides the computational complexity in the
 context of a classical UTM, and its hardware implementation that gives the computational complexity within the UMM paradigm, and show that this algorithm requires only a linear number of interacting memprocessors.
\bigskip
\subsubsection{Algorithm for the solution of the SSP}\label{intro_fourier_SSM_sec}  ~\smallskip\\
Let us consider a set of integer numbers $G=\{a_1,\dots,a_n\}$ and the function
\begin{equation}
g(x)=-1+\prod_{j=1}^{n}(1+e^{i2\pi a_{j}x}).\label{product}
\end{equation}
By expanding the product in~(\ref{product}) it is then obvious that we can store all possible $2^n-1$ products
\begin{equation}
\prod_{j\in P}e^{i2\pi a_{j}x}=\exp\left[i2\pi x\sum_{j\in P}a_{j}\right],\label{exp}
\end{equation}
where $P$ is a set of indexes taken from all possible different non-empty subsets of $\{1,\dots,n\}$. This implies that the function $g(x)$ contains information on {\it all} sums of {\it all} possible sub-sets of $G$.

Let us now consider the discrete Fourier transform (DFT)
\begin{equation}
F(f_{h})=\mathcal{F}\{g(x)\}=\frac{1}{N}\sum_{k=1}^{N}g(x_{k})e^{i2\pi f_{h}x_{k}}.
\label{FFT}
\end{equation}

If this DFT has a sufficient number of points, it will show a peak in correspondence of each $f_{h}$, with magnitude equal to the number of sub-sets of $G$ that sum exactly to $f_{h}$.
\bigskip
\subsubsection{Numerical implementation}\label{numeric_sec}~\smallskip\\
The first step for an efficient and accurate solution of \eqref{FFT} requires the determination of the maximum frequency $f_{\max}$ such that $F(f>f_{\max} )$ and $F(f<-f_{\max})$ are negligible \cite{wiley_enc}. This maximum frequency can be easily determined in our case. By defining $G_{+}$
the sub-set of positive elements of $G$, and $G_{-}$ that of negative elements, we have
\begin{equation}
f_{\max}=\max\left\{  \sum_{j\in G_{+}}a_{j},-\sum_{j\in G_{-}}a_{j}\right\},
\end{equation}
which can be approximated in excess as
\begin{equation}
f_{\max}<n\max\{\left\vert a_{j}\right\vert \}.
\end{equation}
We note again that the transform of $g(x)$ will show peaks in correspondence to the natural numbers in between $-f_{\max}$ and $f_{\max}$. Since the elements of $G$ are integers, $g(x)$ is a periodic function with period $T$ at most equal to 1. In this case, we can then apply the discrete fast Fourier transform (FFT) which, according to the theory of DFTs and the sampling theorem \cite{wiley_enc}, provides the {\it exact} spectrum of $g(x)$. Indeed, from the theory of harmonic balance \cite{wiley_enc}, we can define a number of points
\begin{equation}
N=2f_{\max}+1,
\end{equation}
and divide the period $T$ in sub-intervals of amplitude $\Delta x=N^{-1}$, namely
\begin{equation}
x_{k}=\frac{k}{N}\text{ \ \ with }k=0,...,N-1,
\end{equation}
and then obtain the DFT of $g(x_{k})$ using the FFT algorithm.

In order to determine the complexity of our numerical algorithm let us indicate with $n$ the number of input elements and with $p$ the precision of the problem, namely the number of binary values that it takes to state the problem. In our case then, $n$ is the cardinality of $G$, while $p$ is proportional to the number of discretization points $N$. In order to determine $g(x)$ for every $x_k$ we then need to compute $np$ complex exponentials and $np$ multiplications of complex variables. Therefore, a total of $4np$ floating-point operations.

Instead, in order to determine the DFT we need to make the following important observation. If we want to determine only one sum $f_{h}=s$, we can use Goertzel's algorithm \cite{Goertzel} which is linear in $p$. Therefore, the algorithm to solve the SSP for a given sum $s$ requires $O(np)$ operations: it is linear in both $n$ and $p$. However, we point out that $p$ is not bounded by $n$ and it depends on $N$, which scales exponentially with the number of bits used to represent $f_{\max}$.

\begin{figure}
\centerline{
\includegraphics[width=1\columnwidth]{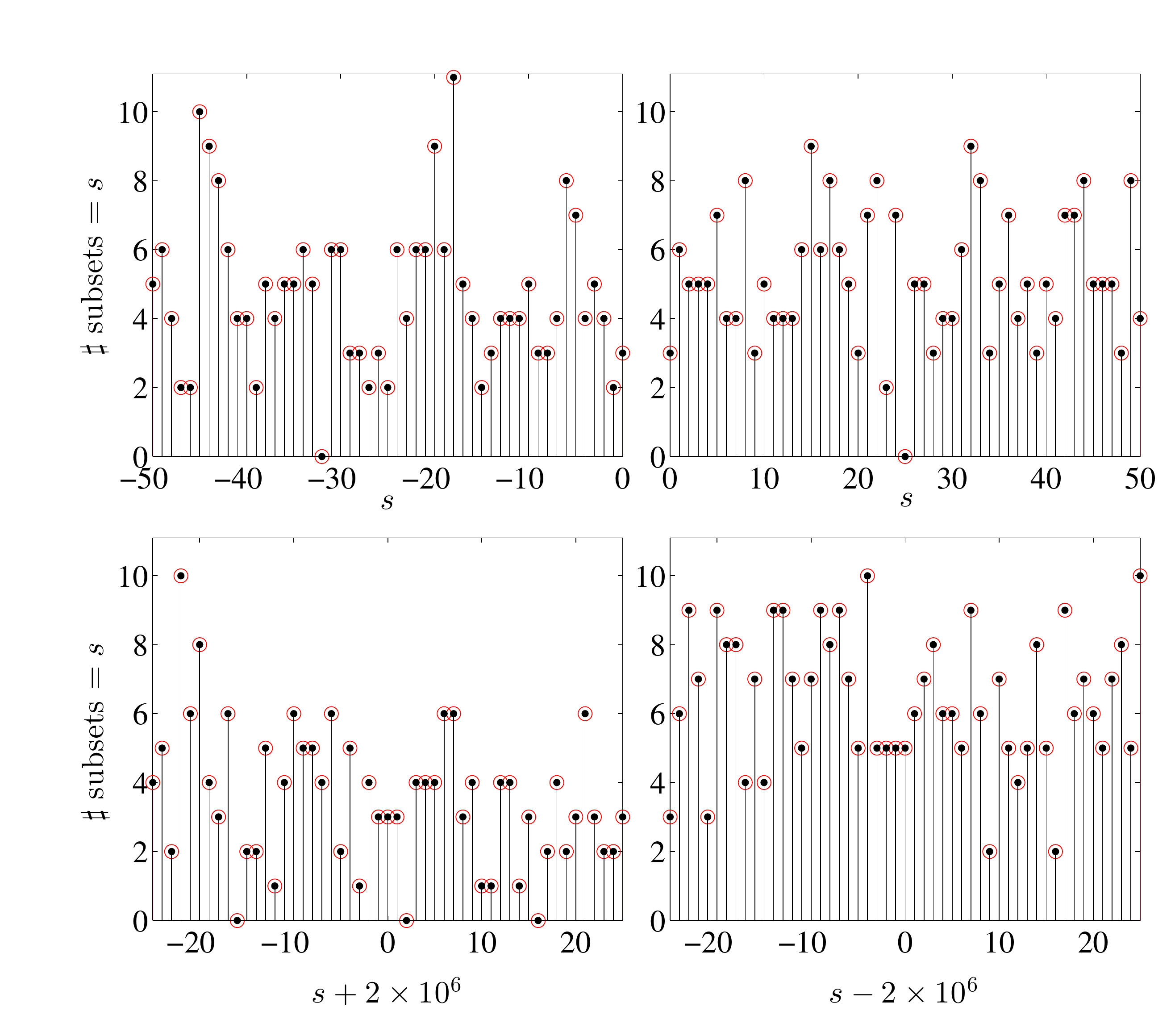}}
\caption{\label{SSP_fourier}Solution of the subset-sum problem using our proposed FFT-based method (black dots) and dynamic programming (red circles). The set $G$ has been randomly chosen and contains the following integers: $G=\{$$-3,639,314;$ $3,692,922;$ $797,045;$ $498,601;$ $-3,550,452;$ $3,530,307;$ $1,220,548;$ $-1,490,478;$ $132,492;$ $-981,923;$ $-4,240,333;$ $-2,600,841;$ $-3,766,812;$ $-3,160,924;$ $-2,600,478;$ $-827,335;$ $-4,503,456;$ $4,027,146;$ $4,447,855;$ $-91,368;$ $-107,483;$ $-1,622,812;$ $4,000,519;$ $-1,307,540;$ $-3,887,975;$ $2,802,502;$ $-1,102,621$$\}$. The figure shows
only a few representative solutions (of all possible solutions found with both methods) in the neighborhood of $s=0$ and close to the edges of the maximum frequency.}
\end{figure}

On the other hand, if we want to compute the solution for all $s$ values simultaneously we need to use the FFT which scales as $O(p\log(p))$. Therefore, the final order of complexity is $O((n+\log(p))p)$. Note that the best algorithm for large $p$ belongs to the class of ``dynamic programming'' algorithms which are order $O(np)$ \cite{algorithms_book}. Therefore, our numerical method has the same complexity as the best known algorithm. However, even if the computational complexity class is the same, for practical Turing computation dynamic programming is better because it involves only sums of integers and boolean logic. On the other hand, our algorithm is extremely important because it can be directly implemented in memcomputing architectures reducing the UMM complexity to $O(n)$ as we show in the next section. Finally, in figure \ref{SSP_fourier} we show a characteristic sample of solutions of the SSP obtained with the proposed method and the dynamic programming for a set $G$ containing 27 elements.

\bigskip
\subsubsection{Hardware implementation}\label{hardware_implementation}~\smallskip\\

\begin{figure}
\centerline{
\includegraphics[width=1\columnwidth]{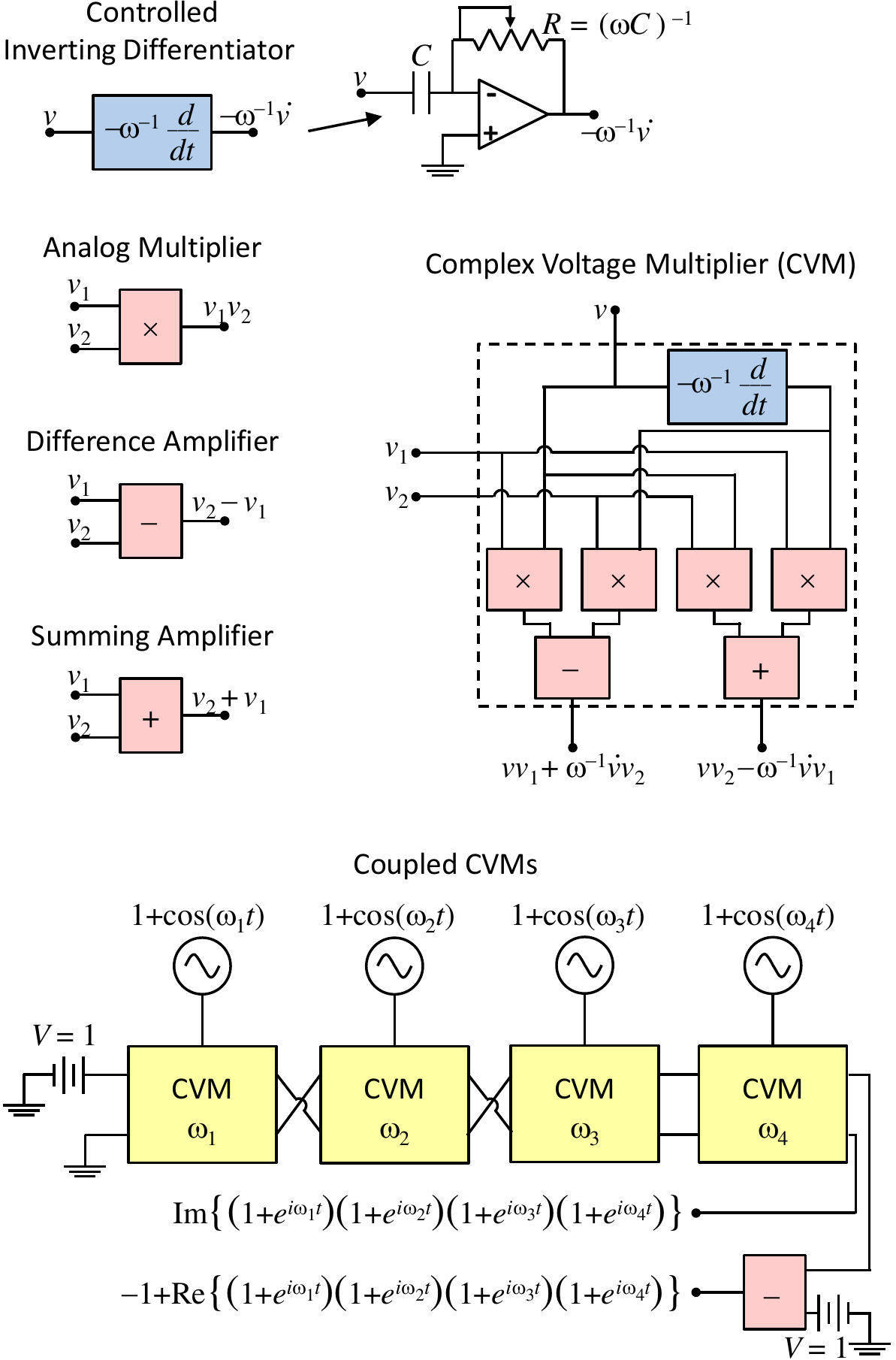}}
\caption{\label{CVM}System of four connected memprocessors. Each memprocessor is composed of a complex voltage multiplier (CVM) and a sinusoidal voltage source. The CVM is composed of standard analog operational circuits. In particular the inverting differentiator employs a variable resistance in order to obtain a desired constant multiplying the derivative. The data is encoded and stored in each memprocessor by setting the frequency of the generator: at the frequency $\omega_j$ corresponds the datum $a_j=\omega_j/2\pi$. Moreover, the sign of $a_j$ can be set by inverting both the input terminals and the output terminals, as shown in the the second CVM of the figure. The connected memory cells have two output states that are the real and imaginary parts of the function $g(t)=-1+\prod_j(1+\exp[i\omega_jt])$, where each $\omega_j$ can be set positive or negative.}
\end{figure}

\begin{figure}
\centerline{
\includegraphics[width=1\columnwidth]{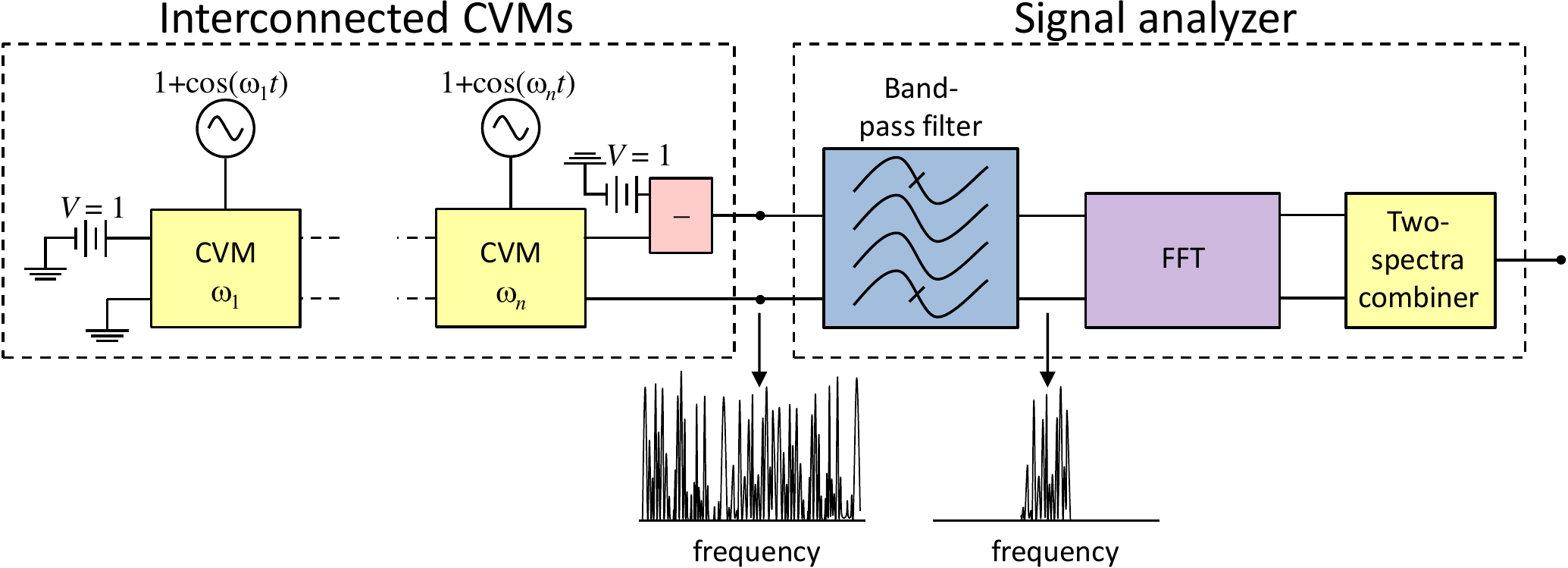}}
\caption{\label{band_pass}Signal analyzer used as read-out unit to select only a window of frequency amplitudes without increasing the computational burden.}
\end{figure}

Before describing the actual hardware implementation of the method, we briefly set a connection between the exponential information overhead of section \ref{exp_IO_sec} and the general framework of the method in section \ref{intro_fourier_SSM_sec}.

We consider memprocessors with infinite discrete states belonging to $M=M_a$. The first step is to notice that the state vector $u_j$ associated with one memprocessor can be taken as the vector of the amplitudes of one of the products in \eqref{product}, thus $u_j$ has $u_{j_0}=1$ and $u_{j_{a_j}}=1$, while all the other components are 0. Then, the operation defined in section \ref{exp_IO_sec} reduces to the standard sum and multiplication, in particular $\star$ and $\oplus$ are sums and $\ast$ is a multiplication. Thus, these simple relations prove that there is an ideal system of interconnected memprocessors that can solve the SSP by using only a linear number of memprocessors exploiting the exponential information overhead.

In addition to this, we can also devise a system of interconnected memprocessors--that can be easily fabricated in the laboratory--that works exactly as we want. In order to prove this claim we define a complex voltage multiplier (CVM) as described in figure \ref{CVM}. This is a two-port device, the first and second ports work as the \lq\lq{}in\rq\rq{} and \lq\lq{}out\rq\rq{} terminals of the memprocessor described in section \ref{exp_IO_sec}. The CVM can be easily built using standard amplifiers as described in figure \ref{CVM}. Each CVM is connected to a generator that applies a voltage $v_j=1+\cos(\omega_jt)$, where the frequencies $\omega_j$ are related to the elements of $G$ through $\omega_j=2\pi a_j$. The state of a single isolated CVM-memprocessor can be read by connecting the first port to a DC generator of voltage $V=1$ and the other port to a signal analyzer that implements in hardware the FFT. On the other hand, by connecting the CVM-memprocessors as in figure \ref{CVM}, and using a signal analyzer at the last port as in figure \ref{band_pass}, we can directly read the solution of the SSP.

We make now two important remarks on the complexity of this approach. The first concerns the execution time. Indeed, in the numerical implementation, the period $T$ of the function \eqref{product} is only a mathematical entity, and what matters is the number of points $N$ used to sample $T$. On the contrary, in the hardware implementation, what matters for the execution time (excluding the FFT) to obtain \eqref{product} is only $T$, which is bounded because it is $1$ (in the appropriate unit), so that it is independent of both $n$ and $p$. Secondly, in order to perform the FFT with a signal analyzer we would in principle need a number of time points of the order of $N$ evaluated as in section \ref{numeric_sec}. This would be a problem even for a signal analyzer because $N$ grows unbounded. However, there is a straightforward solution to this (generally embedded directly in the signal analyzers as depicted in Figure~\ref{band_pass}): before connecting the signal analyzer we interpose a band-pass filter to select only a range of frequencies so the time samples needed by the signal analyzer will be only a bounded number independent of both $n$ and $p$. Therefore, this approach solves the SSP in just {\it one step} for {\it any} $s\in\mathbb{Z}$, when implemented in hardware.

Finally, it is worth noticing that we have the solution of the {\it decision} version of the SSP problem, i.e., we only know whether there is a subset that sums to $s$, but we do not know which actual subset is the solution. Knowing the particular subset that sums to $s$ is the {\it optimization} version of the SSP and is not NP-complete but NP-hard (that is actually harder to solve!) \cite{complexity_bible}.

However, it is easy to show that the hardware implementation we propose is able to solve also the the latter version of the SSP in polynomial time. In fact, to find a particular subset that sums to $s$ we can read the frequency spectrum around $s$ for different configurations of the machine: in the \nth{1} configuration we use all CVMs turned on. In the \nth{2} configuration we set $\omega_1=0$. If the amplitude corresponding to $s$ is $>1$ we let $\omega_1=0$ for the next configuration, otherwise we turn on again $\omega_1$ and $a_1=\omega_1/(2\pi)$ is an element of the subset we are searching for. In the \nth{3} configuration we repeat the same procedure of the \nth{2} but with $\omega_2$. By iterating this procedure for a number of times $\leq n$, we then find one of the subsets (if not unique) that sums to $s$.

\section{Conclusions}\label{Conclusions}
In summary, we have introduced the concept of {\it universal memcomputing machines}, as a class of general-purpose machines that employ memory elements to both store information and process it (memprocessors). The UMMs have unique features that set them apart from Turing machines and which mirror those of the brain: they boast {\it intrinsic parallelism} (their transition functions act simultaneously on all memprocessors), {\it functional polymorphism} (there is no need to change topology to compute different functions, just the input signals), and {\it information overhead} (physically interconnected memprocessors allow for storing and retrieval of a quantity of information which is not simply proportional to the number of memory cells). These last two features, in particular, are unique to UMMs and cannot be found in any other machine that has been suggested in the past, like, e.g., the liquid machine. In addition, although all these features follow from the definition of UMMs they are quite distinct properties.

These general features allow UMMs to solve NP--complete problems in polynomial time with or without the need of an exponentially increasing number
of memprocessors, depending on whether information overhead is used (without) or not (with). Although we have not proved that they can solve some (Turing) undecidable problems, such as the halting problem, they represent a powerful computing paradigm that if realized in practice can overcome many limitations of our present-day computing platforms such as the von Neumann bottleneck and the ever-increasing energy requirements of Turing-based machines.

Indeed, a practical implementation of a UMM can already be accomplished by using memelements such as memristors, memcapacitors or meminductors, although the concept can be implemented with any system with memory, whether passive or active. For instance, in this work we have proposed a simple topologically-specific architecture that, if realized in hardware, can solve the subset-sum problem in just one step with a linear
number of memprocessors.

It is finally worth mentioning that although we have taken inspiration from the brain to formalize the concept and properties of universal memcomputing machines, we expect their practical realization to shed valuable light on the operation of the brain itself. We therefore hope our work will motivate both theoretical and experimental studies aimed at developing memcomputing machines with computational power and features that are getting tantalizing closer to those of the brain.

\section{Acknowledgments}
We thank L. Thiele, Y.V. Pershin and F. Bonani for useful discussions. This work has been partially supported by the Center for Magnetic Recording Research at UCSD.

\bibliographystyle{ieeetr}
\bibliography{UMM}

\end{document}